\documentclass[nohyperref]{article}
\usepackage[utf8]{inputenc}
\usepackage{microtype}
\usepackage{graphicx}
\usepackage{subfigure}
\usepackage{booktabs} %

\usepackage{hyperref}
\usepackage{multirow}
\usepackage{verbatim}
\usepackage{soul}
\usepackage{tikz}
\usepackage{tikz-qtree}

\usepackage{amsmath,amsfonts,bm}

\def\eqref#1{eq.~\ref{#1}}
\def\1{\bm{1}}

\DeclareMathAlphabet{\mathsfit}{\encodingdefault}{\sfdefault}{m}{sl}
\SetMathAlphabet{\mathsfit}{bold}{\encodingdefault}{\sfdefault}{bx}{n}

\def\gL{{\mathcal{L}}}

\def\gS{{\mathcal{S}}}

\def\gZ{{\mathcal{Z}}}

\newcommand{\E}{\mathbb{E}}

\newcommand{\R}{\mathbb{R}}

\newcommand{\KL}{D_{\mathrm{KL}}}

\usepackage[accepted]{sty/icml2023}

\usepackage{amsmath}
\usepackage{amssymb}
\usepackage{mathtools}
\usepackage{amsthm}
\usepackage{enumitem}
\usepackage{newtxmath}
\usepackage{xcolor}

\definecolor{orchid}{HTML}{AF72B0}
\definecolor{teal}{HTML}{51a1ba}

\newcommand{\kolya}[1]{\textcolor{blue}{KM: }}
\newcommand{\yoshua}[1]{\textcolor{red}{YB: }}
\newcommand{\edward}[1]{\textcolor{orchid}{EH: }}
\newcommand{\moksh}[1]{\textcolor{orange}{MJ: }}
\newcommand{\katie}[1]{\textcolor{teal}{KE: }}
\newcommand{\alex}[1]{\textcolor{green}{AG: }}

\usepackage[capitalize,noabbrev]{cleveref}

\makeatletter
\renewcommand*{\sectionautorefname}{\S\@gobble}
\renewcommand*{\subsectionautorefname}{\S\@gobble}
\renewcommand*{\subsubsectionautorefname}{\S\@gobble}
\makeatother



\theoremstyle{plain}

\theoremstyle{definition}

\theoremstyle{remark}

\usepackage[textsize=tiny]{todonotes}
\icmltitlerunning{GFlowNet-EM for Learning Compositional Latent Variable Models}

\begin{document}

\twocolumn[
\icmltitle{GFlowNet-EM for Learning Compositional Latent Variable Models}
\icmlsetsymbol{equal}{*}

\begin{icmlauthorlist}
\icmlauthor{Edward J. Hu}{equal,milaudem}
\icmlauthor{Nikolay Malkin}{equal,milaudem}
\icmlauthor{Moksh Jain}{milaudem}
\icmlauthor{Katie Everett}{gr,mit}
\icmlauthor{Alexandros Graikos}{sb}
\icmlauthor{Yoshua Bengio}{milaudem,cifar}

\end{icmlauthorlist}

\icmlaffiliation{milaudem}{Mila, Universit\'e de Montr\'eal}
\icmlaffiliation{gr}{Google Research}
\icmlaffiliation{sb}{Stony Brook University}
\icmlaffiliation{mit}{Massachusetts Institute of Technology}
\icmlaffiliation{cifar}{CIFAR Fellow}

\icmlcorrespondingauthor{Edward J.\ Hu}{edward@edwardjhu.com}

\icmlkeywords{Machine Learning, ICML}

\vskip 0.3in
]

\printAffiliationsAndNotice{\icmlEqualContribution} %

\begin{abstract}
Latent variable models (LVMs) with discrete compositional latents are an important but challenging setting due to a combinatorially large number of possible configurations of the latents. A key tradeoff in modeling the posteriors over latents is between expressivity and tractable optimization. For algorithms based on expectation-maximization (EM), the E-step is often intractable without restrictive approximations to the posterior. We propose the use of GFlowNets, algorithms for sampling from an unnormalized density by learning a stochastic policy for sequential construction of samples, for this intractable E-step. By training GFlowNets to sample from the posterior over latents, we take advantage of their strengths as amortized variational inference algorithms for complex distributions over discrete structures. Our approach, GFlowNet-EM, enables the training of expressive LVMs with discrete compositional latents, as shown by experiments on non-context-free grammar induction and on images using discrete variational autoencoders (VAEs) without conditional independence enforced in the encoder.

Code: \mbox{ \href{https://github.com/GFNOrg/GFlowNet-EM}{github.com/GFNOrg/GFlowNet-EM}.}
\end{abstract}

\section{Introduction}

In the real world, we often observe high-dimensional data that is generated from lower-dimensional latent variables~\cite{Bishop2007PatternRA}. In particular, it is often natural for these latent variables to have a discrete, compositional structure for data domains like images and language. For example, an image might be decomposed into individual objects that have a relationship between their positions, %
and natural language utterances contain individual words that describe relationships between abstract concepts.
Modeling this discrete compositional latent structure allows for combining existing concepts in new ways, an important inductive bias for human-like generalization~\citep{goyal2022inductive}.

\begin{figure}[t]
\hspace{-0.0\linewidth}\includegraphics[width=1.\linewidth,trim=0 10 0 0]{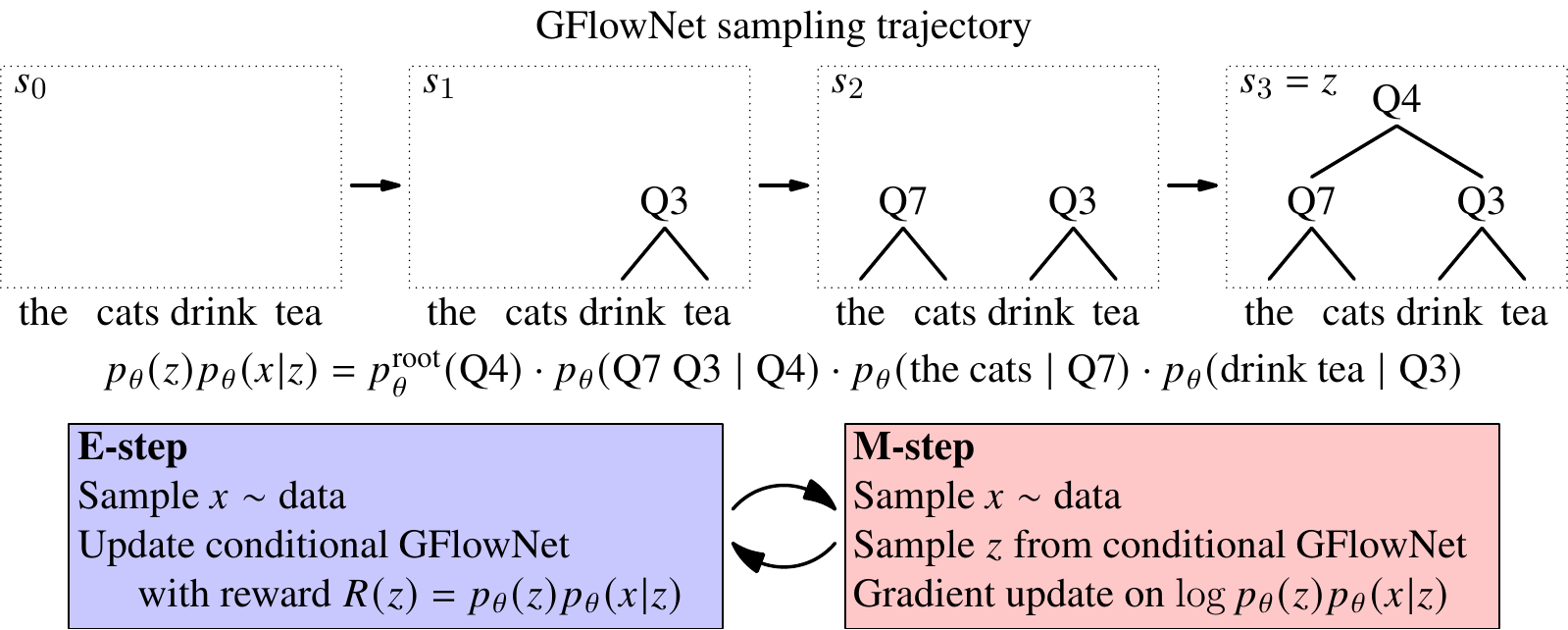}\vskip-0.25em
\caption{GFlowNet-EM for training a latent variable model $p_\theta(z)p_\theta(x|z)$ to maximize likelihood of observed data $x$. The generative model here is a probabilistic context-free grammar. The GFlowNet samples a latent parse tree $z$ from an approximation to the posterior $p_\theta(z|x)$. GFlowNet-EM can flexibly handle non-context-free grammars, black-box priors on tree shape, etc. (\autoref{sec:exp-grammars}).}
 \label{fig:figure_one}
\end{figure}

One family of approaches for maximum-likelihood estimation in LVMs is based on the expectation-maximization algorithm \citep[EM;][]{dempster1977em}, which we review in~\autoref{sec:background-em}.
However, inference of the posterior over latent variables, which is needed in the E-step of EM, is generally intractable when there are combinatorially large number of possible configurations for the latents, such as when the latent random variable does not factorize and represents a discrete compositional structure like a tree or graph.
One can approximately sample from this posterior by running Markov Chain Monte Carlo (MCMC), which can be prohibitively expensive and suffer from poor mixing properties.
Another approach is to impose conditional independence assumptions on the generative model or on the posterior approximation; the latter is known as variational EM (see \autoref{sec:background-em}).
Both limit the expressivity of the LVM.
One such example studied here is the induction of context-free grammars~\cite{baker1979trainable}, which has a generative model under which the expansion of a symbol is independent of its context.

Generative flow networks \citep[GFlowNets;][]{bengio2021flow,bengio2021foundations}, which we review in~\autoref{sec:background-gfn}, are an amortized inference method for sampling from unnormalized densities by sequentially constructing samples using a learned stochastic policy. This sequential construction makes GFlowNets especially useful for sampling discrete compositional objects like trees or graphs.
In this work, we propose to use GFlowNets to learn an amortized sampler of the intractable posterior conditioned on a data sample (\autoref{fig:figure_one}).
This enables the learning of LVMs without conditional independence assumptions, or with weaker ones compared to traditional LVMs like probabilistic context-free grammars (PCFGs).
We also make several algorithmic contributions to mitigate the optimization challenges in jointly learning a GFlowNet sampler and a generative model, notably, posterior collapse~\cite{wang2021posterior}, when the learned posterior only models a few of the modes of the true posterior.
We validate our method, which we call GFlowNet-EM, on both language and image domains.
We intend for this work to serve as a tool for learning more powerful latent variable models that were previously prohibitively expensive to learn.

Our contributions include:
\begin{enumerate}[left=0pt,nosep,label=(\arabic*)]
    \item The GFlowNet-EM framework for maximum likelihood estimation in discrete compositional LVMs that are intractable to optimize by exact EM;
    \item Algorithmic improvements to stabilize joint learning with the generative model while mitigating posterior collapse;
    \item Empirical demonstrations of LVMs with intractable posteriors learned with GFlowNet-EM, including a non-context-free grammar and a discrete VAE without independence assumptions in the encoder. 
\end{enumerate}

\section{Background}

\subsection{Expectation-Maximization (EM)}
\label{sec:background-em}
We review the standard formulation of the EM algorithm \citep{dempster1977em} and its variational form~\citep{neal1998view,koller2009probabilistic}. Consider a LVM with a directed graphical model structured as $z\to x$, with likelihood given by $p(x)=\sum_zp_\theta(z)p_\theta(x|z)$. The latent $z$ may itself have hierarchical structure and be generated through a sequence of intermediate latent variables. Given a dataset $\{x^i\}_{i=1}^T$, we wish to optimize the parameters $\theta$ to maximize the data log-likelihood
\begin{equation}
\gL=\log\prod_{i=1}^Tp(x^i)=\sum_{i=1}^T\log\sum_zp_\theta(z)p_\theta(x^i|z).
\label{eq:ll}
\end{equation}
The EM algorithm achieves this by maximizing a variational bound on~\autoref{eq:ll}, known as the evidence lower bound (ELBO) or negative free energy:
\begin{align}\vspace{-1mm}
\gL
&\geq\sum_{i=1}^T\E_{z\sim q(z|x^i)}\log \frac{p_\theta(z)p_\theta(x^i|z)}{q(z|x^i)}\nonumber\\
&=\gL-\sum_{i=1}^T \KL(q(z|x^i)\|p(z|x^i)),\label{eq:elbo}
\end{align}
where $p(z|x^i)\propto p_\theta(z)p_\theta(x^i|z)$ is the true posterior over the latent. The inequality holds for any collection of distributions $q(z|x^i)$ and is an equality if and only if $q$ equals the true posterior. 

An important choice in EM algorithms is how to parameterize and store the distributions $q(z|x^i)$. In simple EM applications like mixture models, they are stored in a tabular way, i.e., as a matrix of logits that represents the true posterior (\emph{exact EM}). In other settings, $q$ is constrained to lie in a simpler family of distributions, and this family need not contain the true posterior (\emph{variational EM}). A common simplifying assumption is one of conditional independence between components of $z$, e.g., if $z=(z_1,z_2,z_3)$, then $q(z|x^i)=q(z_1|x^i)q(z_2|x^i)q(z_3|x^i)$ (see \autoref{sec:factorization_handicaps_learning}). Finally, in \emph{amortized variational EM}, $q(z|x^i)$ can be parametrized as a neural network, as we will describe below.

The EM algorithm iterates two steps, each of which increases the ELBO (\autoref{eq:elbo}):

\paragraph{E-step.} Optimize the distributions $q(z|x^i)$ so as to approximately make $q(z|x^i)\propto p_\theta(z)p_\theta(x^i|z)$. If $q$, or its factors, are stored in a tabular way, this step is as simple as appropriately normalizing the full matrix $p_\theta(z)p_\theta(x^i|z)$. In other applications, such as for fitting VAEs, $q$ can be optimized using gradient steps to minimize $\KL(q(z|x^i)\|p(z|x^i))$.

\paragraph{M-step.} Optimize $\gL$ with respect to the parameters of $p$, as by taking gradient steps on 
\vspace*{-2mm}\begin{equation}\vspace*{-2mm}
\E_{i}[\E_{z\sim q(z|x^i)}\log p_\theta(z)p_\theta(x^i|z)].
\label{eq:m_step}
\end{equation}

\paragraph{Amortized variational EM.}

In amortized variational EM, $q$ is parametrized by a neural network $q_\phi$ taking $x^i$ as input, which allows evaluation of $q_\phi(z|x)$ at any $x$ and thus generalization to unseen data: sampling from $q(-|x^i)$ becomes easy,  
at the amortized cost of having to train the neural net.
The ELBO can also be jointly optimized with respect to the parameters of both $q$ and $p$ instead of through separate E and M steps.
This is the principle behind VAE models~\citep{JimenezRezende2014StochasticBA,Kingma2014AutoEncodingVB}. 

\paragraph{Wake-sleep for EM.} We return to the question of the E-step -- optimizing $q$ -- when $q$ is parametrized as a neural network $q_\phi(z|x)$. To maximize the ELBO (\autoref{eq:elbo}), $q$ needs to be trained to minimize $\KL(q(z|x^i)\|p(z|x^i))$ for data samples $x^i$.\footnote{Such training can not be done directly in general, since the true posterior is unknown, but algorithms, including GFlowNet-EM, use the fact that $p(z|x^i)\propto p_\theta(z)p_\theta(x^i|z)$, which is available.} If $z$ is high-dimensional, this network can be difficult to train and $q_\phi$ may not assign high likelihood to all modes of the true posterior (\emph{posterior collapse}): when a mode is not represented in $q$, no sample from that mode is ever drawn, which would make it impossible to update $q$ to represent that mode. Instead, $q$ tends to focus on a single mode, even if it can in principle represent multiple modes.

The \emph{sleep phase}, a procedure originally used for fitting posteriors over latents in deep stochastic networks \citep{Hinton1995TheA} but later generalized to other settings \citep{bornschein2015rws,Le2019RevisitingRW,Hewitt2020LearningTL}, aims to mitigate posterior collapse. In the sleep phase, latents $z\sim p_\theta(z)$ and data $x\sim p_\theta(x|z)$ are hallucinated from the generative model (`dreamt', as opposed to `wakeful' use of real data $x^i$), and $q_\phi(z|x)$ is optimized with respect to its likelihood of recovering $z$. That is, the objective minimizes
\vspace*{-2mm}\begin{equation}\vspace*{-2mm}
\E_{z\sim p_\theta(z),x\sim p_\theta(x|z)}[-\log q_\phi(z|x)].
\label{eq:sleep_phase}
\end{equation}
For a given $x$, this objective is equivalent to minimizing $\KL(p_\theta(z|x)\|q_\phi(z|x))$, the opposite direction of the KL compared to~\autoref{eq:elbo}. This direction of the KL will cause $q_\phi$ to seek a broad approximation to the true posterior that captures all of its modes, preventing posterior collapse. On the other hand, if hallucinated samples $x$ are not close to the distribution of the real data $x^i$, the sleep phase may not provide a useful gradient signal for the posteriors $q_\phi(z|x^i)$ that are used in the M-step~\autoref{eq:m_step} with real $x^i$. Therefore, both wake and sleep E-steps can be combined in practice \citep{bornschein2015rws,Le2019RevisitingRW}.

\subsection{GFlowNets}
\label{sec:background-gfn}

We briefly review GFlowNets and their training objectives. For a broader introduction, the reader is directed to \citet{malkin2022trajectory}, whose conventions and notation we borrow, and to other papers listed in~\autoref{sec:related-gfn}.

GFlowNets \citep{bengio2021flow} are a family of algorithms for training a stochastic policy to sample objects from a target distribution over a set of objects $\gZ$ (such as complete parse trees, in \autoref{fig:figure_one}). The set $\gZ$ is a subset of a larger \emph{state space} $\gS$, which contains partially constructed objects (like the incomplete parse trees in the first three panels of \autoref{fig:figure_one}). Formally, the state space has the structure of a directed acyclic graph, where vertices are \emph{states} and edges are \emph{actions} that transition from one state to another. There is a designated \emph{initial state} $s_0$ with no parents (incoming edges), while the \emph{terminal states} -- those with no children (outgoing edges) -- are in bijection with the complete objects $\gZ$. A \emph{complete trajectory} is a  sequence of states $s_0\rightarrow s_1\rightarrow\dots\rightarrow s_n=z$, where $x\in\gZ$ and each $s_i\rightarrow s_{i+1}$ is an action (like an addition of a node to the parse tree).

A \emph{(forward) policy} is a collection of distributions $P_F(s'|s)$ over the children of every nonterminal state $s\in\gS\setminus \gZ$. A policy induces a distribution over complete trajectories $\tau=(s_0\rightarrow\dots\rightarrow s_n)$ given by $P_F(\tau)=\prod_{i=1}^nP_F(s_i|s_{i-1})$. This distribution can be sampled by starting at $s_0$ and sequentially sampling actions from $P_F$ to reach the next state. The policy $P_F$ also induces a distribution $P_F^\top$ over the terminal states via
\begin{equation}
P_F^\top(z)=\sum_{\text{$\tau$ leading to $z$}}P_F(\tau).
\label{eq:pfterm}
\end{equation}
That is, $P_F^\top(z)$ is the marginal likelihood that a trajectory sampled from $P_F$ terminates at $z$.

\paragraph{Training GFlowNets.} Given a reward function $R:\gZ\to\R_{\geq0}$, the goal of GFlowNets is to learn a parametric policy $P_F(s'|s;\theta)$ such that $P_F^\top(z)\propto R(z)$, i.e., the policy samples an object with likelihood proportional to its reward. Because $P_F^\top$ is a (possibly intractable) sum over trajectories (\ref{eq:pfterm}), auxiliary quantities need to be introduced to optimize for reward-proportional sampling. The most commonly used objective in recent work, trajectory balance \citep[TB;][]{malkin2022trajectory}, requires learning two models in addition to the forward policy: a \emph{backward policy} $P_B(s|s';\theta)$, which is a distribution over the \emph{parents} of every noninitial state, and a scalar $Z_\theta$, which is an estimate of the partition function (total reward). The TB objective for a trajectory $\tau=(s_0\rightarrow\dots\rightarrow s_n=z)$ is 
\vspace*{-2mm}\begin{equation}\vspace*{-2mm}
\gL_{\rm TB}(\tau;\theta)=\left[\log\frac{Z_\theta\prod_{i=1}^nP_F(s_i|s_{i-1};\theta)}{R(z)\prod_{i=1}^nP_B(s_{i-1}|s_i;\theta)}\right]^2.
\end{equation}
If this loss is made equal to 0 for all trajectories $\tau$, then the policy $P_F(-|-)$ samples proportionally to the reward. (From now on, we omit the dependence of $P_F$, $P_B$, and $Z$ on $\theta$ for simplicity.)

In practice, this loss can be minimized by gradient descent on $\theta$ for trajectories sampled either \emph{on-policy}, taking $\tau\sim P_F(\tau)$ from the current version of the policy, or \emph{off-policy}. Just as in reinforcement learning (RL), off-policy training can be done in various ways, such as by sampling $\tau$ from a tempered version $P_F^\#$ of the current policy or by sampling $\tau\sim P_B(\tau|z)$ from the \emph{backward} policy starting at a known terminal state. \citet{madan2022learning} introduce subtrajectory balance (SubTB), which generalizes TB to partial trajectories. 

\paragraph{Conditional GFlowNets.} GFlowNets can be conditioned on other variables \citep{bengio2021foundations,jain2022multiobjective,robust-scheduling}. If the reward depends on a variable $x$, then the learned models $P_F$, $P_B$, and $Z$ can all take $x$ as an input and be trained to sample from the conditional reward $R(z|x)$. GFlowNet-EM makes critical use of this ability to model the posterior conditioned on a given data sample.

\section{Motivating Example: Pitfalls of Factorization}
\label{sec:factorization_handicaps_learning}

To illustrate the drawbacks of a factorized posterior, we consider a hierarchical version of a Gaussian mixture model as a toy example. The data is generated from a set of superclusters, in which each supercluster has a set of subclusters, which we call `petals' because each is located at a fixed offset around the supercluster mean as in~\autoref{fig:figure_three}. The data generation process first selects which supercluster, then which petal subcluster, a point should be sampled from, and then samples the point from a standard normal distribution centered at the component mean that is determined by the supercluster mean $\mu_i$ plus the appropriate offset for the selected petal $j$. This problem illustrates a setting where the true posterior $p(i,j|x)$ has a dependence between the discrete latent factors $i$ and $j$, where $i$ denotes the supercluster and $j$ denotes the petal subcluster.

We consider a small version of this problem with four supercluster means arranged in a grid shape where each supercluster has four petals. We use a fixed variance and uniform priors over the choice of supercluster and petal for each data point. The model must learn only the positions of the supercluster means so as to maximize the data likelihood.

\begin{figure}[t]
\centering
\hspace{-0.0\linewidth}
\begin{tabular}{@{}c@{}c@{}c@{}}
Exact & Factorized & GFlowNet\\
\includegraphics[width=0.33\linewidth,trim=0 10 0 0]{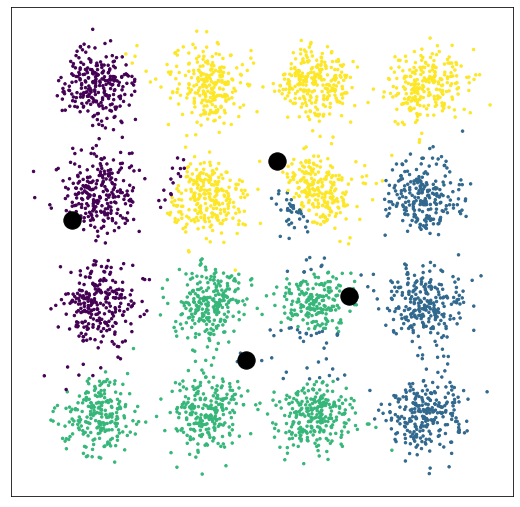}
&
\includegraphics[width=0.33\linewidth,trim=0 10 0 0]{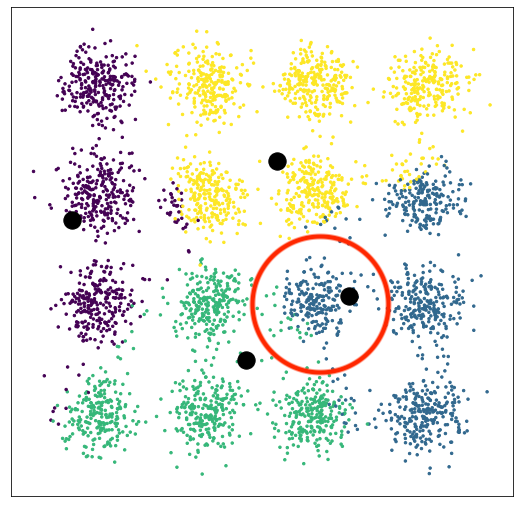}
&
\includegraphics[width=0.33\linewidth,trim=0 10 0 0]{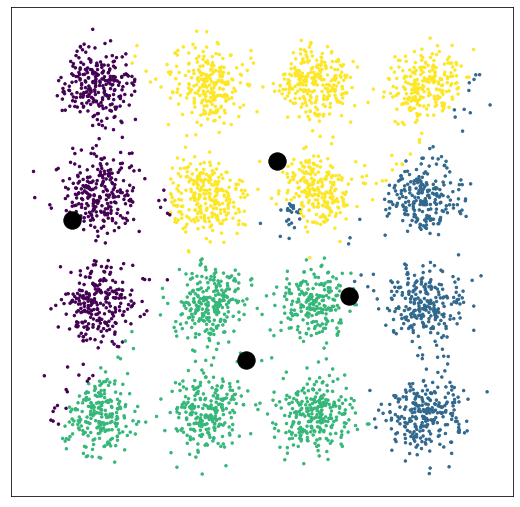}
\\
\includegraphics[width=0.33\linewidth,trim=0 10 0 0]{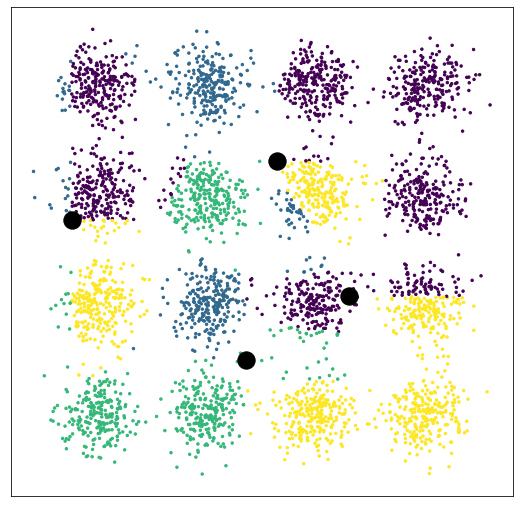}
&
\includegraphics[width=0.33\linewidth,trim=0 10 0 0]{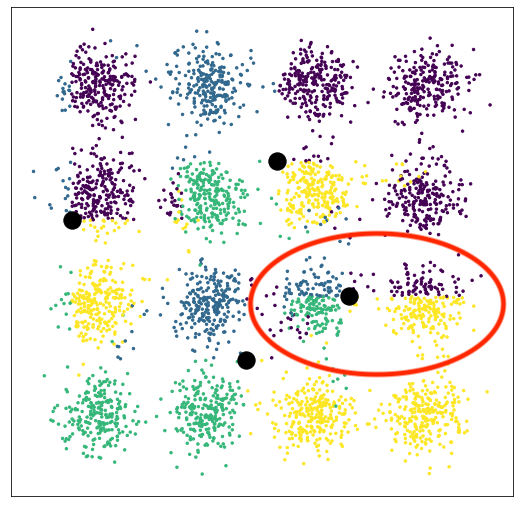}
&
\includegraphics[width=0.33\linewidth,trim=0 10 0 0]{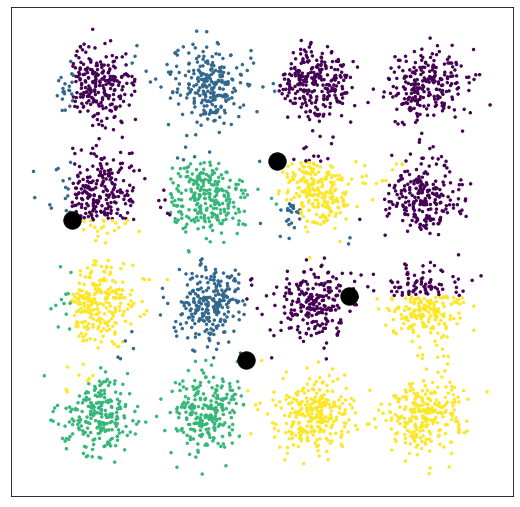}
\end{tabular}
\caption{Posteriors $q(i,j|x)$ inferred during a single E-step for a particular estimate of supercluster means (black dots). \textbf{Top row:} colour indicates which \emph{supercluster} (value of $i$) each point is assigned to. \textbf{Bottom row:} colour indicates which \emph{petal} (value of $j$) each point is assigned to. Assignments use the most likely pair $(i,j)$ for each point in the posterior $q(i,j|x)$. Note the different behaviour of the factorized posterior in the areas circled in red.}
 \label{fig:figure_two}
\end{figure}

This arrangement induces multiple modes in the true posterior $p(i,j|x)$ for a particular estimate of the supercluster means $\mu$; for example, there can be ambiguity about whether a certain point came from the top left petal of one supercluster or the top right petal of another supercluster (\autoref{fig:figure_two}). This requires the inference algorithm to perform combinatorial reasoning to infer optimal assignments (i.e., considering all $(i,j)$ combinations), which is a notoriously difficult problem for algorithms that use a mean-field posterior approximation. 

In this problem, we can easily perform the exact E-step by modeling the posterior in a tabular fashion, where $q(i,j|x)$ is computed exactly as a categorical distribution over all possible pairs $(i,j)$. However, if we were to increase the number of levels of the hierarchy, with each point explained by a combination of many more than two factors, computing the exact posterior would become intractable. 

Meanwhile, \emph{factorized} posteriors can be computed analytically for generative models with this structure \citep{ghahramani1994factorial}. To alleviate the scalability limitations as the depth of the hierarchy grows, we could perform variational EM using the mean-field assumption, so that the approximate posterior is factorized as $q(i,j|x) = q(i|x)q(j|x)$ and a separate categorical distribution is modeled over each latent factor. Yet, as seen in \autoref{fig:figure_two} the factorized approximation fails to assign the proper posterior, and as shown with~\autoref{fig:figure_three}, EM with a factorized approximation to the posterior fails to recover the true supercluster means even on this small dataset. This simple example illustrates the fundamental limitations of factorized posteriors. In more complicated problems, e.g., models for layer separation in computer vision \citep{foreground-background-em}, this effect can become more pronounced.

In contrast, the posterior learned by GFlowNet-EM, which makes no independence assumptions on the approximate posterior, achieves a better fit to the true posterior while being more scalable. We elaborate on this approach in the next section.

\section{GFlowNet-EM}

\begin{algorithm}[t]
  \caption{GFlowNet-EM: Basic form with thresholding}
  \label{alg:gfnem}
\begin{algorithmic}[1]
\REQUIRE Data $\{x^i\}$, generative model with parameters $\theta$, GFlowNet with parameters $\phi$, optimization and exploration hyperparameters, threshold $\alpha$
\REPEAT
\STATE Sample $x^i\sim{\rm data}$
\STATE Sample $\tau\sim P_F^\#(\tau|x^i)$;  $z\gets\text{(last state of $\tau$)}$
\STATE $\gL\gets\text{[TB loss along $\tau$ with reward $p_\theta(z)p_\theta(x|z)$]}$
\STATE E-step: gradient update on $\phi$ with $\nabla_\phi\gL$
\IF{$\gL<\alpha$}  
\STATE Sample $\tau\sim P_F(\tau|x^i)$;  $z\gets\text{(last state of $\tau$)}$
\STATE M-step: gradient update on $\theta$ with $\nabla_\theta[-\log p_\theta(z)p_\theta(x|z)]$ %
\ENDIF
\UNTIL some convergence condition
\end{algorithmic}
\end{algorithm}

The GFlowNet-EM algorithm simultaneously trains two models: the generative model $p_\theta(z,x)$, factorized as $p_\theta(z)p_\theta(x|z)$, and a conditional GFlowNet $q(z|x)=P_F^T(z|x)$ that approximates the true posterior $p_\theta(z|x)$. %

\paragraph{E-step.} The GFlowNet is conditioned on $x$ and trained to sample $z$ with reward $R(z|x)=p_\theta(z)p_\theta(x|z)$. If trained perfectly, the GFlowNet's marginal terminating distribution $P_F^\top(z|x)$ -- note the dependence of $P_F$ on the conditioning variable $x$ -- is proportional to $R(z|x)$, and thus the policy $P_F(-|-,x)$ samples from the true posterior.

In the problems we study, $z$ is a discrete compositional object, and a state space needs to be designed to enable sequential construction of $z$ by a GFlowNet policy. We describe the state space for each setting in our experiments in the corresponding section (\cref{sec:experiments}).

\paragraph{M-step.} The terminating distribution $P_F^\top(z|x)$ of the GFlowNet is used as a variational approximation to the posterior to perform updates to the generative model's parameters. Namely, for a data sample $x^i$, we sample a terminal state -- a latent $z$ -- from the policy of the conditional GFlowNet and perform a gradient update on $\log p_\theta(z)p_\theta(x^i|z)$, thus performing in expectation a gradient update on (\ref{eq:m_step}).

Note that because the generative model $p_\theta$ evolves over the course of joint optimization, the reward for the GFlowNet is nonstationary. E-steps and M-steps are alternated in the course of training, and the schedule of gradient updates -- number of GFlowNet updates in between successive M-steps -- is a parameter that can be fixed or chosen adaptively. We discuss the challenges arising from joint training, and solutions to them, in \cref{sec:gfnem-tricks}.

The basic form of the algorithm, including an adaptive E-step schedule, is presented as \cref{alg:gfnem}.

\subsection{GFlowNet-EM Optimization Techniques}
\label{sec:gfnem-tricks}

GFlowNet-EM presents two challenges that are not present in standard GFlowNet training. First, the estimated posterior $q(z|x)$ is conditioned on the data point $x$, and the dependence of the reward function on $x$ may be complex. Second, the GFlowNet is trained with a nonstationary reward, as the generative model $p$, which provides the reward, changes over the course of GFlowNet-EM training. On the other hand, it is important for the GFlowNet to track the true posterior as it evolves, so as not to bias the M-step and produce degenerate solutions.
We employ a variety of new and existing techniques to address these two challenges. The existing techniques are reviewed in~\autoref{sec:tricks_appendix}. Ablation studies are presented in~\autoref{app:grammar-ablation} and~\autoref{sec:vae} to demonstrate the effectiveness of individual techniques.

\paragraph{Adaptive E-steps via loss thresholding.}%
If the GFlowNet were able to model the true posterior perfectly, one could reduce the GFlowNet loss to zero after every M-step (yielding exact EM).
This is, however, unrealistic due to finite model capacity and compute constraints.
We propose a method for adaptively choosing the number of updates to the GFlowNet that are performed in between successive M-step gradient updates.
Treating a moving average of the GFlowNet's training loss as an indicator of how well the true posterior is approximated, we heuristically set a loss \emph{threshold}, and perform an M-step gradient update after an update to the GFlowNet only if this moving average falls below the threshold. A lower threshold corresponds to requiring a more accurate approximate posterior for updating the generative model. Because the posterior tends to become simpler to model during the course of training from a random initialization, we use a heuristic threshold schedule that linearly decreases the requisite threshold to trigger an M-step update.

\paragraph{Local credit assignment with modular log-likelihood.}
In some interesting LVMs, such as those in \S\ref{sec:exp-grammars}, the reward decomposes as a product of terms accumulated over steps of the sampling sequence. In this case, a forward-looking SubTB loss as described in \citet{additive-energies} can be used as the GFlowNet objective instead of TB.

\paragraph{Exploratory training policy.}
Off-policy exploration in GFlowNet training can be used to improve mode coverage.
The ability of GFlowNets to be stably trained off-policy is a
key strength compared to other variational inference algorithms \citep{malkin2022gfnhvi}. As described in~\autoref{sec:experiments}, we employ two exploration methods: \emph{policy tempering} (making $P_F^\#(s'|s,x)$ proportional to $P_F(s'|s,x)^\beta$ for some $\beta<1$) and \emph{$\epsilon$-uniform sampling} (making $P_F^\#(s'|s,x)$ a mixture of $P_F(s'|s,x)$ and a uniform distribution over the action space).

\subsection{Improving Posterior Estimation}

\paragraph{A sleep phase for GFlowNet-EM.}

\begin{algorithm}[t]
  \caption{GFlowNet-EM: E-step (sleep phase)}
  \label{alg:sleep_phase}
\begin{algorithmic}[1]
\STATE Sample $z\sim p_\theta(z),x\sim p_\theta(x|z)$
\STATE Sample trajectory $\tau\sim P_B(\tau|z,x)$ leading to $z$
\STATE Gradient step on $\phi$ with $\nabla_\phi[-\log P_F(\tau|x)]$
\end{algorithmic}
\end{algorithm}

We propose adding a sleep phase to the E-step updates of GFlowNet-EM, taking advantage of the ability to sample ancestrally from the generative model to prevent posterior collapse.

The sleep phase requires minimizing $-\log q(z|x)$ as in~\autoref{eq:sleep_phase} for $z,x$ sampled ancestrally from the generative model.
However, $q(z|x)=P_F^\top(z|x)$ is a (possibly intractable) sum of likelihoods of all sampling trajectories leading to $z$.
To optimize this log-likelihood, we sample a trajectory leading to $z$ from the \emph{backward} policy $\tau\sim P_B(\tau|z,x)$ and optimize the parameters of the \emph{forward} policy $P_F$ with objective $-\log P_F(\tau|x)$.
This amounts to maximizing the log-likelihood that the GFlowNet's sampling policy conditioned on $x$ recovers $z$ by following the sampling trajectory $\tau$.
It can be shown that for any fixed value of the parameters of $P_B$, the global optimum of this objective with respect to $P_F$ is a maximizer of $\log P_F^\top(z|x)$, guaranteeing correctness. 
Theoretical results and experiments related to this maximum likelihood training objective for GFlowNets can be found in \citet{zhang2023unifying}.

\paragraph{MCMC using GFlowNet as the proposal distribution.}

Another way to leverage the generative model to better estimate the posterior is to run a short MCMC chain initialized with samples drawn from the GFlowNet to bring them closer to the true posterior distribution. The MCMC proposal can make use of the GFlowNet policy itself, using the `back-and-forth' proposal of~\citet{zhang2022generative}.

\section{Empirical Results}
\label{sec:experiments}

\subsection{Hierarchical Mixture Revisited}
\label{sec:hierarchical_mixture_results}

As our first experiment, we compare exact EM, variational EM with a factorized posterior, and GFlowNet-EM on the hierarchical mixture dataset presented in~\autoref{sec:factorization_handicaps_learning}. For GFlowNet-EM, the E-step is performed by a GFlowNet conditioned on the data. The GFlowNet's policy, parametrized as a small MLP, takes two actions: the first action chooses the supercluster assignment and the second action chooses the petal assignment. The reward can be set to $R(i,j|x)=p(x|i,j)$, proportional to the posterior $p(i,j|x)$ as the prior $p(i,j)$ is uniform.

\begin{figure}[t]
\centering
\hspace{-0.05\linewidth}
\begin{tabular}{@{}c@{}c@{}c@{}}
\textbf{Exact EM} & \textbf{Variational EM} & \textbf{GFlowNet-EM}\\
\includegraphics[width=0.33\linewidth,trim=0 10 0 0]{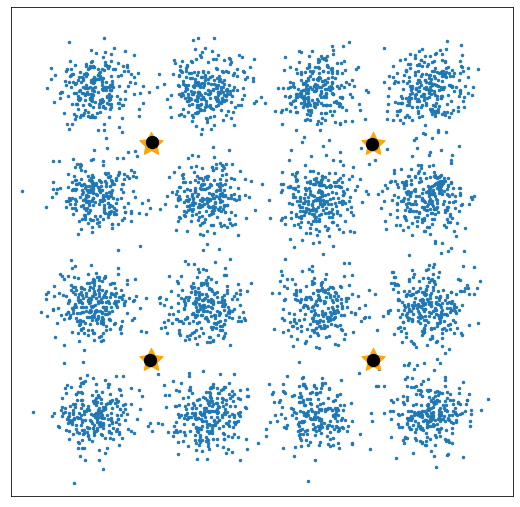}
&
\includegraphics[width=0.33\linewidth,trim=0 10 0 0]{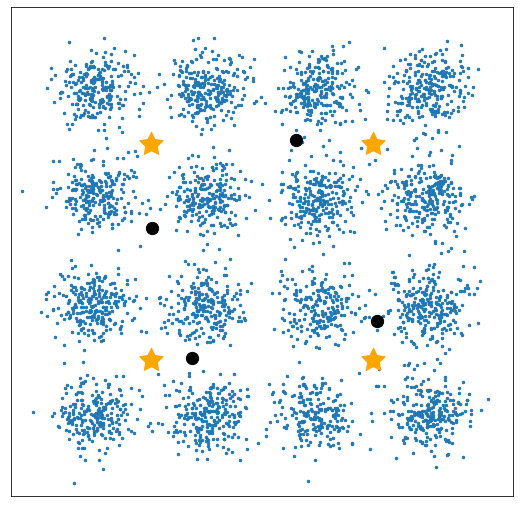}
&
\includegraphics[width=0.33\linewidth,trim=0 10 0 0]{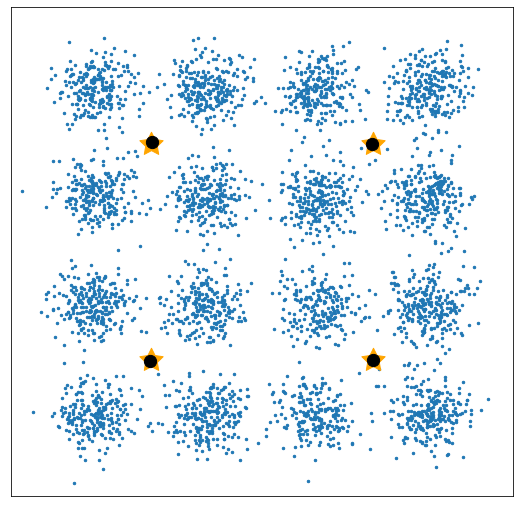}
\end{tabular}
\vspace{-0.05\linewidth}
\caption{Estimated supercluster means are shown as black dots while ground truth supercluster means are shown as orange stars. Unlike the (factorized) Variational EM, GFlowNet-EM puts the means at the right place.}
 \label{fig:figure_three}
\end{figure}

Averaged over twenty random seeds, after sixty iterations (which induces convergence in all methods), the data log-likelihood per sample for exact EM is $-5.79 \pm 0.74$, variational EM is $-7.26 \pm 1.12$, and GFlowNet-EM is $-5.77 \pm 0.48$. For reference, the average log-likelihood for the ground truth supercluster means, used to sample the dataset, is $-5.62 \pm 0.01$. Implementation details are described in \autoref{sec:hierarchical_appendix}. The estimated supercluster means for each method on a single initialization are shown in \autoref{fig:figure_three}, where exact EM and GFlowNet-EM both nearly match the ground truth supercluster means but variational EM fails to learn the correct means.

\subsection{Grammar Induction on Penn Tree Bank (PTB)}
\label{sec:exp-grammars}

In linguistics and in the theory of formal languages, a grammar refers to a set of structure constraints on sequences of symbols.
Since \citet{chomsky}, all dominant theories have assumed some form of hierarchical generative grammar as a universal feature of natural languages.
The task of grammar induction asks whether one can automatically discover from data the \emph{hierarchical} grammar that explains the \emph{sequential} structure we observe, and whether the discovered rules coincide with ones created by human experts.
We study the case with binary rule branching.
See \autoref{app:grammar-setup} for a more detailed description of the assumptions we make on the grammar and the way the rule likelihoods are parametrized.

\paragraph{Dataset.}
We use a subset of Penn Tree Bank \citep[PTB;][]{marcus1999treebank} that contains sentences with 20 or fewer tokens.
Otherwise, we follow the preprocessing done by \citet{kim-etal-2019-compound}, including removing punctuation and tokenizing OOV words.
The vocabulary size (number of T symbols) is 9672.
We use 30 NT symbols and 60 PT symbols.

\paragraph{Baselines.}
We reproduce the Neural PCFG architecture from~\citet{kim-etal-2019-compound}.
Taking advantage of specialized algorithms for context-free grammars, we either marginalize over the latent space (\textit{Marginalization}\footnote{The marginal likelihood has the same gradient as exact sampling EM in expectation.}) or sample from the true posterior (\textit{Exact sampling EM}). 
Our \textit{Marginalization} baseline matches the result produced by the public repository of~\citet{kim-etal-2019-compound}.
The Monte-Carlo EM (MC-EM) baseline draws samples from the posterior by running 1000 MCMC steps with a proposal distribution that performs random single tree rotations and symbol changes.
All baseline and GFlowNet-EM runs are run for 10,000 grammar (M-step) gradient updates.
We use Torch-Struct~\cite{rush-2020-torch} to perform marginalization and exact sampling in PCFGs.

\paragraph{Metrics.}
We use two metrics to evaluate learned grammars: 
\begin{enumerate}[label=(\arabic*),nosep,left=0pt]
\item The marginal likelihood of a held-out dataset under the learned grammar, which can be equivalently expressed in terms of negative log-likelihood per word.
When marginalization is not tractable, we use a variational upper bound described in~\autoref{app:grammar-ncfg}. %
\item How well the parse trees under the learned grammar resemble human-labeled trees, as measured by an F1 score between sets of spans (constituents) in a proposed and a human-labeled parse tree, following \citet{kim-etal-2019-compound}. This metric evaluates the linguistic relevance of the learned grammar.
\end{enumerate}

\paragraph{GFlowNet-EM parametrization.}
The GFlowNet models the posterior over possible parse trees given a sentence (a sequence of Ts, i.e., terminal symbols).
Even though we only consider binary trees, following~\citet{kim-etal-2019-compound}, the number of possible trees is exponential both in the sequence length and in the number of PTs and NTs.\footnote{Even the number of binary tree \emph{shapes} for a sequence of length $n$ is the Catalan number $\frac{1}{n}\binom{2(n-1)}{n-1}=O(4^n/n\sqrt n)$.}
We propose a bottom-up GFlowNet action space, which incrementally joins two adjacent partial trees by hanging them under a common parent, as illustrated in \autoref{fig:figure_one}.
The initial state is represented by the sequence of $n$ terminal symbols $x$, each of which is a tree of depth zero.
A binary parse tree is obtained after $n-1$ joining steps.
We only generate the NT symbols in the tree and marginalize over PT symbols, as this can be done in linear time (see \autoref{app:ptm}).
We use a Transformer~\cite{vaswani17attn} with full attention over root nodes and a bottom-up MLP aggregator; see~\autoref{app:grammar-parametrization} for more details and \autoref{app:grammar-ablation} for ablations studying the different components of GFlowNet-EM. 

In addition to the basic algorithm in~\autoref{alg:gfnem}, we use a forward-looking SubTB loss, a sleep phase~(\autoref{alg:sleep_phase}), and MCMC steps as described in~\autoref{sec:gfnem-tricks} for the grammar induction experiment.

\begin{table}[t]
\vspace*{-4mm}
\centering
\caption{Inducing a context-free grammar (CFG) or a non-context-free-grammar (Non-CFG) using different methods. GFlowNet-EM allows the incorporation of an energy-based model (EBM) prior or the use of an intractable grammar, e.g., Non-CFG. All configuration are run over 5 random seeds.}
\resizebox{1\linewidth}{!}{
\begin{tabular}{@{}llcc}
\toprule
Grammar &    Method         & NLL / word $\downarrow$ &  Sentence F1 $\uparrow$ \\ \midrule
 \multirow{6}{*}{CFG}    & Marginalization    & $5.61\pm 0.01$ & $39.51\pm7.01$\\
\cmidrule(lr){2-4}
 & Exact-sampling EM   & $5.74\pm 0.05$ &  $31.17\pm6.06$ \\\cmidrule(lr){2-4} %
     & MC-EM    &  $5.88\pm 0.01$ & $22.31\pm 1.04$ \\
     & \small{\ \ \ \ \ \ + EBM Prior}    & $5.91\pm0.02$ &  $23.81\pm1.41$ \\
     & GFlowNet-EM   & $5.70\pm0.03$ &  $34.85\pm3.39$ \\
     & \small{\ \ \ \ \ \ + EBM Prior}  & $5.79\pm 0.03$ &  ${\bf 48.41\pm 1.38}$\\\midrule
\multirow{2}{*}{Non-CFG} & MC-EM    &  -  &  $18.98\pm0.26$   \\
     & GFlowNet-EM   & ${\bf \leq 5.46\pm 0.07}$    &  $38.68\pm1.90$   \\ \bottomrule
\end{tabular}
}
\label{tab:grammar}
\vspace*{-2mm}
\end{table}

\subsubsection{Context-free Grammar}
\label{sec:exp-grammars-cfg}
We first consider the well-studied problem of inducing a binary branching probabilistic context-free grammar (PCFG), where the rule probabilities are independent of the context.
In this case, the true posterior over parse trees is tractable to sample from or even marginalize over using an algorithm with run time cubic in the sequence length \citep{baker1979trainable}. 
Nonetheless, we validate GFlowNet-EM by comparing it with exact EM, i.e., always sampling from the exact posterior.
As \textit{Exact sampling EM} is equivalent to GFlowNet-EM with the constraint that the GFlowNet is perfectly trained to zero loss on every E-step, the exact sampling baseline gives a rough upper bound on the performance of GFlowNet-EM without additional inductive biases.%

\paragraph{Results.}
\textit{Marginalization} baseline performs the best in terms of both NLL and F1, as shown in \autoref{tab:grammar}, which we attribute to its much lower gradient variance compared to drawing samples from the true posterior.
GFlowNet-EM can match and exceed sampling from the exact posterior on both metrics, despite having to learn an approximate posterior sampler.
It is worth noting that while GFlowNet-EM is not necessary in this scenario, it has an asymptotic computational advantage because it amortizes the cost of inference; see~\autoref{app:grammar-run-time} for more details.

We now consider setups where \textit{Marginalization} and \textit{Exact sampling} are not tractable.

\subsubsection{CFG with Energy-based Model Guidance}
\label{sec:exp-grammars-ebm}
It can be useful to bias learned LVMs to incorporate domain-specific knowledge.
For example, we might want the learned grammar to produce parse trees that have shapes resembling ones provided by human annotators for linguistic interest.
This preference for tree shapes is hard to integrate because it is a global attribute, which violates the strong conditional independence assumptions in CFGs that are required for correctness of exact sampling algorithms.

We train an energy-based model (EBM) on the \emph{shapes} of human-labeled trees to represent black-box domain knowledge.
The EBM's density acts as a prior that is multiplied by the usual GFlowNet reward.
We anneal the temperature of this prior to infinity in 10,000 steps, thus only biasing the beginning (symmetry-breaking) phase of the joint learning process.
See more details in~\autoref{app:ebm-prior}.

\paragraph{Results.}
\autoref{tab:grammar} shows that GFlowNet-EM with the EBM prior can learn grammars that produce trees more similar to human annotation compared to \textit{Exact sampling EM} and even \textit{Marginalization}. We also note that the trees generated have a strong right-branching bias, a well-known feature of English syntax. %

\subsubsection{Non-context-free Grammar}
\label{sec:exp-grammars-ncfg}
The context-free assumption in CFGs makes exact sampling from the posterior tractable.
GFlowNet-EM, however, does not require the true posterior to be tractable, as long as there is underlying structure for amortized learning.
To this end, we experiment with a non-context-free grammar (Non-CFG) that allows a rule probability to depend on the parent of the LHS of the rule (\autoref{app:grammar-ncfg}), for which exact sampling from the posterior over parse trees becomes prohibitively expensive.

\paragraph{Results.}
As shown in \autoref{tab:grammar}, GFlowNet-EM on this Non-CFG yields a grammar that has a significantly lower marginal NLL while having a comparable F1 to \textit{Marginalization} on a CFG, despite drawing finite samples from a learned approximate posterior. This is attributed to the more expressive generative model and to inductive biases of its parametrization: we do not incorporate any external knowledge, e.g., an EBM prior, in this experiment.

\subsection{Discrete Variational Autoencoders}
\label{sec:vae} 

Next, we study the problem of learning deep generative models of images with discrete latent representations. This problem was previously posed under the framework of vector-quantized variational autoencoders \citep[VQ-VAE;][]{oord2017neural}.
VQ-VAEs assume a latent space of the form $\{1,\dots,K\}^n$, where $n$ is the length of the latent vector and $K$ is the number of possible values for each position. However, the VQ-VAE decoder represents each value in $\{1,\dots,K\}$ by its representation vector in a vector space $\R^D$, while the encoder predicts a vector in $\R^D$ and maps it to the value in $\{1,\dots,K\}$ whose representation vector is nearest to the prediction. This manner of passing through a high-dimensional continuous space allows passing approximate gradients from the decoder to the encoder using the straight-through estimator \citep{bengio2013estimating}, but is inherently incapable of learning more than a single-point estimate of the posterior over discrete latents.

\paragraph{GFlowNet encoder.}
We propose to use a GFlowNet as the encoder to learn a policy that sequentially constructs the discrete latent representation of an image (\autoref{fig:vae_policy}). The E-step trains the encoder model to match the posterior distribution over the discrete latents $z$ conditioned on an image $x$, and the M-step trains the decoder to minimize the error in reconstructing $x$ from the latent sampled by the encoder. 

Crucially, this approach does not rely on an approximation of gradients, as the E and M steps are decoupled, and admits an expressive posterior by imposing none of the conditional independence constraints on components of the latent that VAE encoders make. Furthermore, VQ-VAEs assume a uniform prior over the discrete latents $z$. However, GFlowNet-EM enables us to also learn a prior distribution, $p_\theta(z)$, \emph{jointly} with the decoder $p_\theta(x|z)$. This is a clear advantage over VQ-VAEs, which can only learn the prior distribution post-hoc, after the encoder is trained. 

We chose an autogressive encoder: it sequentially constructs the discrete latent by sampling one categorical entry at a time, conditioned on the input image and the previously drawn entries (\autoref{fig:vae_policy}). This reduces the complexity of the encoder network while maintaining an advantage over VQ-VAEs, where the posterior is fully factorized.
For these experiments we modify \autoref{alg:gfnem} by (i) not using an adaptive E-step, but alternately performing 400 E-steps and 400 M-steps and  and (ii) using sleep phase exploration (\autoref{alg:sleep_phase}). 

\paragraph{Training and evaluation.}
To train the encoder and decoder networks, we alternate between E- and M-steps, using 400 gradient updates in each step (see \S\ref{app:vae} for details). We found that this was adequate and no adaptive E-step was needed. During E-steps, we exploit the sleep phase for exploration, where we sample $z$ from either the uniform or learned prior and $x$ from the current $p_\theta(x|z)$. We also observe that convergence is accelerated by training the decoder with samples drawn \textit{greedily} from the learned encoder policy, although this gives a biased objective in the M-step and results in slightly lower test data likelihood. %

\begin{table}[t]
\vspace*{-3mm}
\centering
\caption{GFlowNet-EM achieves lower NLL than VQ-VAE on the static MNIST test set (mean and std.\ over 5 runs; GFlowNet NLL estimated using 5000 importance-weighted samples). \textbf{Bold:} the lowest NLL and all those not significantly higher than it ($p>0.1$ under an unpaired $t$-test).}
\resizebox{1\linewidth}{!}{
\begin{tabular}{@{}lccc}
\toprule
 & \multicolumn{3}{c}{Codebook size} \\
\cmidrule(lr){2-4}
Method & $K=4$ & $K=8$ & $K=10$ \\
\midrule
VQ-VAE & $86.36 \pm 0.14$ & $80.84 \pm 0.39$ & $82.96 \pm 0.38$ \\\midrule
GFlowNet-EM & $74.18 \pm 0.41$ & $\bf 70.74 \pm 0.99$ & $\bf 70.67 \pm 0.72$ \\ %
\small{\ \ + Greedy Decoder Training (GD)} & $76.22 \pm 0.58$ & $72.03 \pm 0.98$ & $72.69 \pm 1.56$ \\ %
\small{\ \ + GD + Jointly Learned Prior} & $78.59 \pm 1.48$ & $\bf 70.84 \pm 1.06$ & $\bf 71.69 \pm 1.90$ \\ %
\bottomrule
\end{tabular}
}
\label{tab:vqvae}
\end{table}

\paragraph{Results and discussion.}
We perform our experiments on the static MNIST dataset \cite{deng2012mnist}, with a $4\times4$ spatial latent representation and using dictionaries of sizes $K\in\{4,8,10\}$ and dimensionality $D=1$. We compare with a VQ-VAE with the same latent representation as a baseline. (For codebook sizes larger than 10, we observed the NLL of the VQ-VAE increase.) %
In Table \ref{tab:vqvae} we show estimated NLL on the test set obtained by the VQ-VAE model and different variations of GFlowNet-EM for all dictionary sizes $K$. In all experiments, GFlowNet-EM performs significantly better than VQ-VAE, which we attribute to the higher expressiveness of the posterior.

Just as for VQ-VAEs, decoded samples with the latent drawn from a uniform prior do not resemble real images. When the prior $p(z)$ is also learned jointly with $p(x|z)$, we achieve similar results to those assuming a uniform prior, but also gain the ability to draw reasonable unconditional samples from the prior (\autoref{fig:vae_prior_samples}).%

We note that the more expressive posterior and lower NLL come with an increased training cost. Sampling from the posterior requires multiple forward passes of the GFlowNet encoder, and performing the E and M steps alternately entails more training iterations than are needed for VQ-VAEs.

\section{Related Work}

\subsection{GFlowNets}
\label{sec:related-gfn}

GFlowNets \citep{bengio2021flow,bengio2021foundations} were first formulated as a reinforcement learning algorithm that generalizes maximum-entropy RL \citep{haarnoja2018soft} to settings with multiple paths to the same state. However, recent papers \citep{malkin2022gfnhvi,zimmermann2022variational,zhang2023unifying} place GFlowNets in the family of variational methods, showing that they are more amenable to stable off-policy training than policy gradient approaches to minimizing divergences between distributions. Applications include biological molecule and sequence design \citep{jain2022biological,jain2022multiobjective}, causal structure learning \citep{deleu2022bayesian,nishikawa2022bayesian}, and robust combinatorial optimization \citep{robust-scheduling}. 
Energy-based GFlowNets \citep{zhang2022generative} solve the related problem of fitting a GFlowNet to a nonstationary reward defined by a generative model from which exact sampling is intractable; however, the updates to the generative model are approximate contrastive divergence steps, and inference over latent variables is not performed. GFlowNets were also used as approximate posteriors for an ELBO maximization in \citet{gflowout}.

\subsection{Latent Variable Models and EM}

Discrete LVMs, prominent before the deep learning revolution, continue to motivate research, including on posterior regularization techniques \citep{ganchev2010posterior}, theoretical properties of EM \citep{neath2013convergence}, augmenting classical latent variable models with distributed neural representations \citep{dieng-etal-2020-topic}, adapting discrete LVMs to deep learning-scale data for robust classification \citep{malkin2020mining}, and amortized inference \citep{agrawal2021amortized}.

\subsection{Applications}

\paragraph{Grammar induction.} The literature on automatic grammar induction, briefly described in~\autoref{sec:exp-grammars}, is most focused on probablistic context-free grammars and their variants, thanks to the efficient learning algorithm introduced in~\citet{baker1979trainable} and~\citet{lari1990estimation}.
Many variants have increased the expressivity of PCFGs without relaxing the context-free assumption, such as~\citet{kim-etal-2019-compound} and~\citet{zhao-titov-2021-empirical}.
While the learning of PCFGs can be accelerated with careful implementations~\cite{yang-etal-2021-pcfgs,rush-2020-torch}, their time complexity remains cubic in the length of the sequences and in the number of NT and PT symbols.
PCFG induction has been applied to character-level multilingual language modeling~\cite{jin-etal-2021-character-based} and music modeling with the addition of continuous symbols~\citep{lieck2021recursive}.

\paragraph{Discrete VAEs.} 
Discrete latent representations, as popularized by VQ-VAEs \cite{oord2017neural}, have been shown to successfully capture both abstract and low-level features \cite{esser2021taming,baevski2020wav2vec,dhariwal2020jukebox}. In comparison to their continuous VAE counterparts \cite{Kingma2014AutoEncodingVB}, discrete latent variable models utilize more efficiently the available latent degrees of freedom due to their inherent ability to ignore imperceptible input details. The main limitations of discrete VAE models arise from their use of a continuous relaxation to allow for backpropagation \cite{ramesh2021zeroshot} and the fundamental limitation of having to learn the prior over the latent variable separately. GFlowNet-EM overcomes both of these limitations.

\section{Conclusions}

We presented a novel method for maximum-likelihood estimation in discrete latent variable models that uses GFlowNets as approximate samplers of the posterior for intractable E-steps. Our experiments on non-context-free grammar induction and discrete image representations -- both settings where the LVM has an intractable posterior without additional independence assumptions -- show that GFlowNet-EM outperforms existing approaches. Future work should broaden the applications of GFlowNet-EM to other compositional latent variable models, particularly those with continuous or hybrid latents.

\section*{Acknowledgements}

The authors thank Matt Hoffman, Tuan Anh Le, Donna Vakalis, and the anonymous ICML reviewers for their comments on drafts of the paper, as well as Nebojsa Jojic, Paul Soulos, and Dinghuai Zhang for some helpful discussions. They are also grateful for the financial support from IBM, Samsung, Microsoft and Google.

\bibliographystyle{sty/icml2023}
\bibliography{anthology,main}

\begin{thebibliography}{54}
\providecommand{\natexlab}[1]{#1}
\providecommand{\url}[1]{\texttt{#1}}
\expandafter\ifx\csname urlstyle\endcsname\relax
  \providecommand{\doi}[1]{doi: #1}\else
  \providecommand{\doi}{doi: \begingroup \urlstyle{rm}\Url}\fi

\bibitem[Agrawal \& Domke(2021)Agrawal and Domke]{agrawal2021amortized}
Agrawal, A. and Domke, J.
\newblock Amortized variational inference for simple hierarchical models.
\newblock \emph{Neural Information Processing Systems (NeurIPS)}, 2021.

\bibitem[Baevski et~al.(2020)Baevski, Zhou, Mohamed, and
  Auli]{baevski2020wav2vec}
Baevski, A., Zhou, Y., Mohamed, A., and Auli, M.
\newblock wav2vec 2.0: {A} framework for self-supervised learning of speech
  representations.
\newblock \emph{Neural Information Processing Systems (NeurIPS)}, 2020.

\bibitem[Baker(1979)]{baker1979trainable}
Baker, J.~K.
\newblock Trainable grammars for speech recognition.
\newblock \emph{The Journal of the Acoustical Society of America}, 65\penalty0
  (S1):\penalty0 S132--S132, 1979.

\bibitem[Bengio et~al.(2021)Bengio, Jain, Korablyov, Precup, and
  Bengio]{bengio2021flow}
Bengio, E., Jain, M., Korablyov, M., Precup, D., and Bengio, Y.
\newblock Flow network based generative models for non-iterative diverse
  candidate generation.
\newblock \emph{Neural Information Processing Systems (NeurIPS)}, 2021.

\bibitem[Bengio et~al.(2013)Bengio, L{\'{e}}onard, and
  Courville]{bengio2013estimating}
Bengio, Y., L{\'{e}}onard, N., and Courville, A.~C.
\newblock Estimating or propagating gradients through stochastic neurons for
  conditional computation.
\newblock \emph{arXiv preprint 1308.3432}, 2013.

\bibitem[Bengio et~al.(2023)Bengio, Lahlou, Deleu, Hu, Tiwari, and
  Bengio]{bengio2021foundations}
Bengio, Y., Lahlou, S., Deleu, T., Hu, E., Tiwari, M., and Bengio, E.
\newblock {GFlowNet} foundations.
\newblock \emph{Journal of Machine Learning Research (JMLR)}, 2023.
\newblock To appear.

\bibitem[Bishop(2006)]{Bishop2007PatternRA}
Bishop, C.~M.
\newblock \emph{Pattern Recognition and Machine Learning}.
\newblock Springer, 2006.

\bibitem[Bornschein \& Bengio(2015)Bornschein and Bengio]{bornschein2015rws}
Bornschein, J. and Bengio, Y.
\newblock Reweighted wake-sleep.
\newblock \emph{International Conference on Learning Representations (ICLR)},
  2015.

\bibitem[Chomsky(1965)]{chomsky}
Chomsky, N.
\newblock \emph{Aspects of the Theory of Syntax}.
\newblock MIT Press, 1965.

\bibitem[Deleu et~al.(2022)Deleu, G\'{o}is, Emezue, Rankawat, Lacoste-Julien,
  Bauer, and Bengio]{deleu2022bayesian}
Deleu, T., G\'{o}is, A., Emezue, C., Rankawat, M., Lacoste-Julien, S., Bauer,
  S., and Bengio, Y.
\newblock Bayesian structure learning with generative flow networks.
\newblock \emph{Uncertainty in Artificial Intelligence (UAI)}, 2022.

\bibitem[Dempster et~al.(1977)Dempster, Laird, and Rubin]{dempster1977em}
Dempster, A.~P., Laird, N.~M., and Rubin, D.~B.
\newblock Maximum likelihood from incomplete data via the {EM} algorithm.
\newblock \emph{Journal of the Royal Statistical Society B}, 39\penalty0
  (1):\penalty0 1--38, 1977.

\bibitem[Deng(2012)]{deng2012mnist}
Deng, L.
\newblock The {MNIST} database of handwritten digit images for machine learning
  research.
\newblock \emph{IEEE Signal Processing Magazine}, 29\penalty0 (6):\penalty0
  141--142, 2012.

\bibitem[Dhariwal et~al.(2020)Dhariwal, Jun, Payne, Kim, Radford, and
  Sutskever]{dhariwal2020jukebox}
Dhariwal, P., Jun, H., Payne, C., Kim, J.~W., Radford, A., and Sutskever, I.
\newblock Jukebox: A generative model for music.
\newblock \emph{arXiv preprint 2005.00341}, 2020.

\bibitem[Dieng et~al.(2020)Dieng, Ruiz, and Blei]{dieng-etal-2020-topic}
Dieng, A.~B., Ruiz, F. J.~R., and Blei, D.~M.
\newblock Topic modeling in embedding spaces.
\newblock \emph{Transactions of the Association for Computational Linguistics},
  8:\penalty0 439--453, 2020.
\newblock \doi{10.1162/tacl_a_00325}.
\newblock URL \url{https://aclanthology.org/2020.tacl-1.29}.

\bibitem[Esser et~al.(2021)Esser, Rombach, and Ommer]{esser2021taming}
Esser, P., Rombach, R., and Ommer, B.
\newblock Taming transformers for high-resolution image synthesis.
\newblock \emph{Computer Vision and Pattern Recognition (CVPR)}, 2021.

\bibitem[Frey \& Jojic(2005)Frey and Jojic]{foreground-background-em}
Frey, B. and Jojic, N.
\newblock A comparison of algorithms for inference and learning in
  probabilistic graphical models.
\newblock \emph{IEEE Transactions on Pattern Analysis and Machine
  Intelligence}, 27\penalty0 (9):\penalty0 1392--1416, 2005.

\bibitem[Ganchev et~al.(2010)Ganchev, Gra\c{c}a, Gillenwater, and
  Taskar]{ganchev2010posterior}
Ganchev, K., Gra\c{c}a, J., Gillenwater, J., and Taskar, B.
\newblock Posterior regularization for structured latent variable models.
\newblock \emph{Journal of Machine Learning Research (JMLR)}, 11:\penalty0
  2001–2049, aug 2010.

\bibitem[Ghahramani(1994)]{ghahramani1994factorial}
Ghahramani, Z.
\newblock Factorial learning and the {$EM$} algorithm.
\newblock \emph{Neural Information Processing Systems (NIPS)}, 1994.

\bibitem[Goyal \& Bengio(2022)Goyal and Bengio]{goyal2022inductive}
Goyal, A. and Bengio, Y.
\newblock Inductive biases for deep learning of higher-level cognition.
\newblock \emph{Proceedings of the Royal Society A}, 478\penalty0
  (2266):\penalty0 20210068, 2022.

\bibitem[Haarnoja et~al.(2018)Haarnoja, Zhou, Abbeel, and
  Levine]{haarnoja2018soft}
Haarnoja, T., Zhou, A., Abbeel, P., and Levine, S.
\newblock Soft actor-critic: Off-policy maximum entropy deep reinforcement
  learning with a stochastic actor.
\newblock \emph{International Conference on Machine Learning (ICML)}, 2018.

\bibitem[Hewitt et~al.(2020)Hewitt, Le, and Tenenbaum]{Hewitt2020LearningTL}
Hewitt, L.~B., Le, T.~A., and Tenenbaum, J.~B.
\newblock Learning to learn generative programs with memoised wake-sleep.
\newblock \emph{Uncertainty in Artificial Intelligence (UAI)}, 2020.

\bibitem[Hinton et~al.(1995)Hinton, Dayan, Frey, and Neal]{Hinton1995TheA}
Hinton, G.~E., Dayan, P., Frey, B.~J., and Neal, R.~M.
\newblock The ``wake-sleep'' algorithm for unsupervised neural networks.
\newblock \emph{Science}, 268 5214:\penalty0 1158--61, 1995.

\bibitem[Jain et~al.(2022{\natexlab{a}})Jain, Bengio, Hernandez-Garcia,
  Rector-Brooks, Dossou, Ekbote, Fu, Zhang, Kilgour, Zhang, Simine, Das, and
  Bengio]{jain2022biological}
Jain, M., Bengio, E., Hernandez-Garcia, A., Rector-Brooks, J., Dossou, B.~F.,
  Ekbote, C., Fu, J., Zhang, T., Kilgour, M., Zhang, D., Simine, L., Das, P.,
  and Bengio, Y.
\newblock Biological sequence design with {GFlowNets}.
\newblock \emph{International Conference on Machine Learning (ICML)},
  2022{\natexlab{a}}.

\bibitem[Jain et~al.(2022{\natexlab{b}})Jain, Raparthy, Hernandez-Garcia,
  Rector-Brooks, Bengio, Miret, and Bengio]{jain2022multiobjective}
Jain, M., Raparthy, S.~C., Hernandez-Garcia, A., Rector-Brooks, J., Bengio, Y.,
  Miret, S., and Bengio, E.
\newblock Multi-objective {GFlowNets}.
\newblock \emph{arXiv preprint 2210.12765}, 2022{\natexlab{b}}.

\bibitem[Jin et~al.(2021)Jin, Oh, and Schuler]{jin-etal-2021-character-based}
Jin, L., Oh, B.-D., and Schuler, W.
\newblock Character-based {PCFG} induction for modeling the syntactic
  acquisition of morphologically rich languages.
\newblock In \emph{Findings of the Association for Computational Linguistics:
  EMNLP 2021}, pp.\  4367--4378, Punta Cana, Dominican Republic, November 2021.
  Association for Computational Linguistics.
\newblock \doi{10.18653/v1/2021.findings-emnlp.371}.
\newblock URL \url{https://aclanthology.org/2021.findings-emnlp.371}.

\bibitem[Kim et~al.(2019)Kim, Dyer, and Rush]{kim-etal-2019-compound}
Kim, Y., Dyer, C., and Rush, A.
\newblock Compound probabilistic context-free grammars for grammar induction.
\newblock In \emph{Proceedings of the 57th Annual Meeting of the Association
  for Computational Linguistics}, pp.\  2369--2385, Florence, Italy, July 2019.
  Association for Computational Linguistics.
\newblock \doi{10.18653/v1/P19-1228}.
\newblock URL \url{https://aclanthology.org/P19-1228}.

\bibitem[Kingma \& Welling(2014)Kingma and Welling]{Kingma2014AutoEncodingVB}
Kingma, D.~P. and Welling, M.
\newblock Auto-encoding variational {Bayes}.
\newblock \emph{International Conference on Learning Representations (ICLR)},
  2014.

\bibitem[Koller \& Friedman(2009)Koller and Friedman]{koller2009probabilistic}
Koller, D. and Friedman, N.
\newblock \emph{Probabilistic graphical models: principles and techniques}.
\newblock MIT press, 2009.

\bibitem[Lari \& Young(1990)Lari and Young]{lari1990estimation}
Lari, K. and Young, S.
\newblock The estimation of stochastic context-free grammars using the
  inside-outside algorithm.
\newblock \emph{Computer Speech and Language}, 4\penalty0 (1):\penalty0 35--56,
  1990.

\bibitem[Le et~al.(2019)Le, Kosiorek, Siddharth, Teh, and
  Wood]{Le2019RevisitingRW}
Le, T.~A., Kosiorek, A.~R., Siddharth, N., Teh, Y.~W., and Wood, F.
\newblock Revisiting reweighted wake-sleep for models with stochastic control
  flow.
\newblock \emph{Neural Information Processing Systems (NeurIPS)}, 2019.

\bibitem[Lieck \& Rohrmeier(2021)Lieck and Rohrmeier]{lieck2021recursive}
Lieck, R. and Rohrmeier, M.
\newblock Recursive {Bayesian} networks: Generalising and unifying
  probabilistic context-free grammars and dynamic {Bayesian} networks.
\newblock \emph{Neural Information Processing Systems (NeurIPS)}, 2021.

\bibitem[Liu et~al.(2022)Liu, Jain, Dossou, Shen, Lahlou, Goyal, Malkin,
  Emezue, Zhang, Hassen, Ji, Kawaguchi, and Bengio]{gflowout}
Liu, D., Jain, M., Dossou, B. F.~P., Shen, Q., Lahlou, S., Goyal, A., Malkin,
  N., Emezue, C.~C., Zhang, D., Hassen, N., Ji, X., Kawaguchi, K., and Bengio,
  Y.
\newblock {GFlowOut}: Dropout with generative flow networks.
\newblock \emph{arXiv preprint 2210.12928}, 2022.

\bibitem[Madan et~al.(2023)Madan, Rector-Brooks, Korablyov, Bengio, Jain, Nica,
  Bosc, Bengio, and Malkin]{madan2022learning}
Madan, K., Rector-Brooks, J., Korablyov, M., Bengio, E., Jain, M., Nica, A.,
  Bosc, T., Bengio, Y., and Malkin, N.
\newblock Learning {GFlowNets} from partial episodes for improved convergence
  and stability.
\newblock \emph{International Conference on Machine Learning (ICML)}, 2023.

\bibitem[Malkin et~al.(2020)Malkin, Ortiz, and Jojic]{malkin2020mining}
Malkin, N., Ortiz, A., and Jojic, N.
\newblock Mining self-similarity: Label super-resolution with epitomic
  representations.
\newblock \emph{European Conference on Computer Vision (ECCV)}, 2020.

\bibitem[Malkin et~al.(2022)Malkin, Jain, Bengio, Sun, and
  Bengio]{malkin2022trajectory}
Malkin, N., Jain, M., Bengio, E., Sun, C., and Bengio, Y.
\newblock Trajectory balance: Improved credit assignment in {GFlowNets}.
\newblock \emph{Neural Information Processing Systems (NeurIPS)}, 2022.

\bibitem[Malkin et~al.(2023)Malkin, Lahlou, Deleu, Ji, Hu, Everett, Zhang, and
  Bengio]{malkin2022gfnhvi}
Malkin, N., Lahlou, S., Deleu, T., Ji, X., Hu, E., Everett, K., Zhang, D., and
  Bengio, Y.
\newblock {GFlowNets} and variational inference.
\newblock \emph{International Conference on Learning Representations (ICLR)},
  2023.

\bibitem[Marcus et~al.(1999)Marcus, Santorini, Marcinkiewicz, and
  Taylor]{marcus1999treebank}
Marcus, M.~P., Santorini, B., Marcinkiewicz, M.~A., and Taylor, A.
\newblock Treebank-3.
\newblock \emph{Linguistic Data Consortium, Philadelphia}, 14, 1999.

\bibitem[Neal \& Hinton(1998)Neal and Hinton]{neal1998view}
Neal, R.~M. and Hinton, G.~E.
\newblock A view of the em algorithm that justifies incremental, sparse, and
  other variants.
\newblock In \emph{Learning in graphical models}, pp.\  355--368. Springer,
  1998.

\bibitem[Neath et~al.(2013)]{neath2013convergence}
Neath, R.~C. et~al.
\newblock On convergence properties of the {Monte Carlo EM} algorithm.
\newblock \emph{Advances in modern statistical theory and applications: a
  Festschrift in Honor of Morris L. Eaton}, pp.\  43--62, 2013.

\bibitem[Nishikawa-Toomey et~al.(2022)Nishikawa-Toomey, Deleu, Subramanian,
  Bengio, and Charlin]{nishikawa2022bayesian}
Nishikawa-Toomey, M., Deleu, T., Subramanian, J., Bengio, Y., and Charlin, L.
\newblock Bayesian learning of causal structure and mechanisms with {GFlowNets}
  and variational bayes.
\newblock \emph{arXiv preprint 2211.02763}, 2022.

\bibitem[Pan et~al.(2023)Pan, Malkin, Zhang, and Bengio]{additive-energies}
Pan, L., Malkin, N., Zhang, D., and Bengio, Y.
\newblock Better training of {GFlowNets} with local credit and incomplete
  trajectories.
\newblock \emph{International Conference on Machine Learning (ICML)}, 2023.

\bibitem[Ramesh et~al.(2021)Ramesh, Pavlov, Goh, Gray, Voss, Radford, Chen, and
  Sutskever]{ramesh2021zeroshot}
Ramesh, A., Pavlov, M., Goh, G., Gray, S., Voss, C., Radford, A., Chen, M., and
  Sutskever, I.
\newblock Zero-shot text-to-image generation.
\newblock \emph{International Conference on Machine Learning (ICML)}, 2021.

\bibitem[Rezende et~al.(2014)Rezende, Mohamed, and
  Wierstra]{JimenezRezende2014StochasticBA}
Rezende, D.~J., Mohamed, S., and Wierstra, D.
\newblock Stochastic backpropagation and approximate inference in deep
  generative models.
\newblock \emph{International Conference on Machine Learning (ICML)}, 2014.

\bibitem[Rush(2020)]{rush-2020-torch}
Rush, A.
\newblock Torch-struct: Deep structured prediction library.
\newblock In \emph{Proceedings of the 58th Annual Meeting of the Association
  for Computational Linguistics: System Demonstrations}, pp.\  335--342,
  Online, July 2020. Association for Computational Linguistics.
\newblock \doi{10.18653/v1/2020.acl-demos.38}.
\newblock URL \url{https://aclanthology.org/2020.acl-demos.38}.

\bibitem[van~den Oord et~al.(2016)van~den Oord, Kalchbrenner, and
  Kavukcuoglu]{oord2016pixel}
van~den Oord, A., Kalchbrenner, N., and Kavukcuoglu, K.
\newblock Pixel recurrent neural networks.
\newblock \emph{International Conference on Machine Learning (ICML)}, 2016.

\bibitem[van~den Oord et~al.(2017)van~den Oord, Vinyals, and
  Kavukcuoglu]{oord2017neural}
van~den Oord, A., Vinyals, O., and Kavukcuoglu, K.
\newblock Neural discrete representation learning.
\newblock \emph{Neural Information Processing Systems (NIPS)}, 2017.

\bibitem[Vaswani et~al.(2017)Vaswani, Shazeer, Parmar, Uszkoreit, Jones, Gomez,
  Kaiser, and Polosukhin]{vaswani17attn}
Vaswani, A., Shazeer, N., Parmar, N., Uszkoreit, J., Jones, L., Gomez, A.~N.,
  Kaiser, L., and Polosukhin, I.
\newblock Attention is all you need.
\newblock \emph{Neural Information Processing Systems (NIPS)}, 2017.

\bibitem[Wang et~al.(2021)Wang, Blei, and Cunningham]{wang2021posterior}
Wang, Y., Blei, D.~M., and Cunningham, J.~P.
\newblock Posterior collapse and latent variable non-identifiability.
\newblock \emph{Neural Information Processing Systems (NeurIPS)}, 2021.

\bibitem[Yang et~al.(2021)Yang, Zhao, and Tu]{yang-etal-2021-pcfgs}
Yang, S., Zhao, Y., and Tu, K.
\newblock {PCFG}s can do better: Inducing probabilistic context-free grammars
  with many symbols.
\newblock In \emph{Proceedings of the 2021 Conference of the North American
  Chapter of the Association for Computational Linguistics: Human Language
  Technologies}, pp.\  1487--1498, Online, June 2021. Association for
  Computational Linguistics.
\newblock \doi{10.18653/v1/2021.naacl-main.117}.
\newblock URL \url{https://aclanthology.org/2021.naacl-main.117}.

\bibitem[Zhang et~al.(2022)Zhang, Malkin, Liu, Volokhova, Courville, and
  Bengio]{zhang2022generative}
Zhang, D., Malkin, N., Liu, Z., Volokhova, A., Courville, A., and Bengio, Y.
\newblock Generative flow networks for discrete probabilistic modeling.
\newblock \emph{International Conference on Machine Learning (ICML)}, 2022.

\bibitem[Zhang et~al.(2023{\natexlab{a}})Zhang, Chen, Malkin, and
  Bengio]{zhang2023unifying}
Zhang, D., Chen, R. T.~Q., Malkin, N., and Bengio, Y.
\newblock Unifying generative models with {GFlowNets} and beyond.
\newblock \emph{arXiv preprint 2209.02606v2}, 2023{\natexlab{a}}.

\bibitem[Zhang et~al.(2023{\natexlab{b}})Zhang, Rainone, Peschl, and
  Bondesan]{robust-scheduling}
Zhang, D., Rainone, C., Peschl, M., and Bondesan, R.
\newblock Robust scheduling with {GFlowNets}.
\newblock \emph{International Conference on Learning Representations (ICLR)},
  2023{\natexlab{b}}.

\bibitem[Zhao \& Titov(2021)Zhao and Titov]{zhao-titov-2021-empirical}
Zhao, Y. and Titov, I.
\newblock An empirical study of compound {PCFG}s.
\newblock In \emph{Proceedings of the Second Workshop on Domain Adaptation for
  NLP}, pp.\  166--171, Kyiv, Ukraine, April 2021. Association for
  Computational Linguistics.
\newblock URL \url{https://aclanthology.org/2021.adaptnlp-1.17}.

\bibitem[Zimmermann et~al.(2022)Zimmermann, Lindsten, van~de Meent, and
  Naesseth]{zimmermann2022variational}
Zimmermann, H., Lindsten, F., van~de Meent, J.-W., and Naesseth, C.~A.
\newblock A variational perspective on generative flow networks.
\newblock \emph{arXiv preprint 2210.07992}, 2022.

\end{thebibliography}
\clearpage
\appendix

\onecolumn

\section{On GFlowNet optimization techniques}
\label{sec:tricks_appendix}

For the sake of completeness, we review the SubTB loss from \citet{madan2022learning} and the forward-looking parametrization from \citet{additive-energies}. 

\paragraph{Subtrajectory balance (SubTB).} In addition to the forward and backward policy models $P_F(s'|s)$ and $P_B(s|s')$, one trains a state flow estimator $F(s)$, which outputs a scalar for any state in the GFlowNet state space. If $\tau=(s_0\rightarrow\dots\rightarrow s_n)$ is a complete trajectory and $\tau_{i:j}=(s_i\rightarrow s_{i+1}\rightarrow\dots\rightarrow s_j)$ is its subtrajectory, the SubTB loss $\tau_{i:j}$ is defined as
\[{\cal L}_{\rm SubTB}(\tau_{i:j})=\frac{F(s_i)P_F(s_{i+1}|s_i)P_F(s_{i+2}|s_{i+1}\dots P_F(s_j|s_{i-1})}{F(s_j)P_B(s_i|s_{i+1})P_B(s_{i+1}|s_{i+2})\dots P_B(s_{j-1}|s_j)},\]
where we enforce $F(s)=R(s)$ if $s$ is terminal. Thus ${\cal L}_{\rm SubTB}$ reduces to the TB loss if $s_i$ is initial and $s_j$ is terminal, where $F(s_0)$ is identified with $Z$.

The SubTB objective for a complete trajectory $\tau=(s_0\rightarrow\dots\rightarrow s_n)$ with $\lambda=1$, as defined in \citet{madan2022learning}, is the average of SubTB losses for all of its partial trajectories:
\[{\cal L}(\tau)=\frac{1}{\binom{n+1}{2}}\sum_{0\leq i<j\leq n}{\cal L}_{\rm SubTB}(\tau_{i:j}).\]

\paragraph{Forward-looking loss.} The state flow estimator $F(s)$ used in SubTB is typically parametrized in the log domain, i.e., a neural network taking $s$ as input outputs $\log F(s)$. In the case where one has available a \emph{partial reward} accumulated up to state $s$ -- denoted $\tilde R(s)$ -- the forward-looking parametrization from \citet{additive-energies} parametrizes $\log F(s)=\log\tilde R(s)+g(s;\theta)$, where $g$ is a neural network. 

Note that $\tilde R$ can be an arbitrary estimate of the negative `partial energy' of a state $s$, and $\tilde R\equiv0$ yields the regular SubTB objective. However, natural partial log-rewards exist in cases where the low-reward is close to additive over steps taken in a trajectory. For example, in the case of a parser for a context-free grammar, if $s$ is a partially constructed tree, we take $\log\tilde R(s)$ to be the sum of log-likelihoods under the grammar of the production rules that occur in $s$.

\section{Hierarchical mixture}
\label{sec:hierarchical_appendix}
This section describes the experiment setup for the hierarchical mixture experiments in \autoref{sec:hierarchical_mixture_results}. We sample 20 datasets and one initialization of the supercluster means per dataset, where the initial supercluster means are selected to be random points from the data. For each dataset and initialization, we run exact EM, variational EM, and GFlowNet-EM and compute the log-likelihood under the the estimated supercluster means after the final iteration. We run all methods for 60 iterations, which induces convergence in all methods.

The GFlowNet takes 1000 gradient steps in each E-step, samples one latent assignment per data point from the posterior, and takes one gradient step in each M-step. Note that such an optimization method is chosen simply for illustration; the objective in the M-step is quadratic in the parameters and can be optimized in closed form, a case of a more general algorithm for Gaussian multiple cause models \citep{ghahramani1994factorial}.

We implemented several of the optimization techniques in \autoref{sec:gfnem-tricks} (loss thresholding and epsilon-uniform sampling), but the empirical impacts are negligible on this problem as the basic GFlowNet-EM method easily matches the optimal solution found by exact EM.

\section{Grammar induction}
\label{app:grammar}

\subsection{Experiment setup}
\label{app:grammar-setup}
In our setup, following~\citet{kim-etal-2019-compound}, there are 30 NT symbols and 60 PT symbols in addition to T symbols.
In addition, among these NT symbols is a special root symbol, which we call ROOT, that is fixed to be the root of each parse tree.
For convenience, we shall use NT to refer to non-terminal symbols that are not the special ROOT token.
The root symbol has only one child, which is allowed to be any NT symbols, i.e., ROOT $\rightarrow$ NT.
There are $|NT|$, i.e., number of NT symbols, such rules.
Each NT symbol is allowed to branch in two symbols, each belonging to the union of all NT and PT symbols, i.e., NT $\rightarrow$ \{NT, PT\} \{NT, PT\}.
There are $|NT|^3|PT|^2$ such rules.
Finally, each PT symbol is allowed to turn into a T symbol, i.e., a vocabulary item.
There are $|PT||V|$ such rules of the form PT $\rightarrow$ V, where $|V|$ is the size of the vocabulary.

A naive parametrization would use a table to store individual rule probabilities without assuming any dependencies among them.
This, however, is not conducive to learning expressive and linguistically meaningful grammars as described in~\citet{kim-etal-2019-compound}.
\citet{kim-etal-2019-compound} propose a distributed representation, where each symbol is assigned an embedding and rule probabilities are computed using an MLP. See the Neural PCFG in~\citet{kim-etal-2019-compound} for the specifics of this grammar parametrization.

\subsection{GFlowNet parametrization}
\label{app:grammar-parametrization}
Given an input $x$ represented as a sequence of tokens, we would like to sample a parse tree $z$ according to $p(z|x)$.
We construct the parse tree $z$ bottom-up as illustrated in~\autoref{fig:figure_one}.
The state space of the GFlowNet is the space of ordered forests, where a tree represents a sub-tree in the final $z$.
The action space is all pairs of adjacent trees in an ordered forest.
At every time step, we choose one such pair and join them with a new parent to form a new tree.
We use a Transformer~\cite{vaswani17attn} with full attention, which processes only the root nodes of trees in the ordered forest.
The information from non-root nodes are encoded using an aggregator.
For every binary branching, we recursively compute the embedding of the parent node using the static embedding for the symbol at that node combined with the recursively computed embeddings of its children using an MLP.
The embeddings of root nodes, which now encode whole trees, are passed to the Tranformer encoder.
The policies $P_F$ and $P_B$ and the flow estimator $F$ are implemented as MLP heads on top of the Transformer encoder.
We use a sum-pooling operation for the flow estimator, which gives a scalar for every GFlowNet state regardless of how many trees it contains.

Training hyperparameters are listed in Table~\ref{tab:hyper_grammar}.

\begin{table}[t]
\centering
\caption{Hyperparameters for training the GFlowNet-EM for grammar induction.}
\begin{tabular}{lc}
\toprule
Hyperparameter       & Value                  \\ \midrule
Encoder Transformer: Layers & 6                     \\
Hidden Dimension        & 512 \\
Adam $\beta$            & (0.9, 0.99)            \\
Batch Size           & 32                     \\
$P_F, P_B$: Learning Rate & $10^{-4}$ \\
$Z$: Learning Rate & 0.03\\ 
MCMC Step & 10 \\
Sleep phase weight & 10 \\
SubTB $\lambda$ & 1 \\
EBM Prior temperature start & 1 \\
EBM Prior temperature end & 1000 \\
EBM Prior temperature schedule horizon & 10000 \\
Adaptive threshold max & 6 \\
Adaptive threshold min & 3 \\
Adaptive threshold schedule horizon & 10000 \\
Grammar: MLP Hidden Dimension & 256 \\
Grammar: Learning Rate & 0.001 \\
Grammar: Adam $\beta$ & (0.75, 0.999) \\
\bottomrule
\end{tabular}
\label{tab:hyper_grammar}
\end{table}

\subsection{Marginalizing preterminals}
\label{app:ptm}

The construction of a parse tree $z$ given a sentence $x$ is conventionally done in two steps: 1) tagging and 2) parsing.
The tagging step assigns each terminal (T) symbol a preterminal (PT) symbol, and the parsing steps join nonterminal (NT) symbols and PT symbols alike using NT symbols.
Using a GFlowNet to tag T symbols doubles the number of time steps it needs to construct $z$ for a sentence.
This is especially wasteful considering that tagging step can be easily marginalized over in linear time.
As a result, we use the GFlowNet to join T symbols directly and produce parse trees without any PT symbols.
When evaluating the reward of such trees, we perform a marginalization over all PT symbols in each position, which can be done in linear time for both the context-free grammar and the non-context-free grammar we introduced.

\subsection{Training energy-based model prior}
\label{app:ebm-prior}
In~\autoref{sec:exp-grammars-ebm}, we use an energy-based model (EBM) as a prior on tree shapes.
This EBM is trained with a persistent contrastive divergence objective on gold trees from the training set, where the MCMC proposal used in PCD is a random tree rotation (i.e., replacement of a random subtree of the form $[X [Y Z]]$ by $[[X Y] Z]$ or vice versa) and the buffer reset ratio is 0.1. The EBM architecture was a recursive aggregator similar to that described in \cref{app:grammar-parametrization}: the embedding of a node is a MLP evaluated on a concatenation of the embeddings of its children, the embeddings of leaf nodes are fixed to zero vectors, and the output energy is a pooled embedding of the root. This ensures that the energy depends only on the tree \emph{shape} and not on the symbols. The hyperparameter for training the EBM are listed in \autoref{tab:hyperparams_ebm}. The make the EBM compatible with the Sleep phase, we temper the EBM term in the GFlowNet reward with a schedule that decays it linearly.

\begin{table}[]
\centering
\caption{Hyperparameters for training the EBM prior on tree shapes.}
\begin{tabular}{lc}
\toprule
Hyperparameter       & Value                  \\ \midrule
MLP Hidden Dimension & 16                     \\
Learning Rate        & $10^{-5}$ \\
Adam $\beta$            & (0.9, 0.99)            \\
$L_2$ Regularization    & $10^{-4}$ \\
Batch Size           & 32                     \\
Sequence Length      & 40                     \\ \bottomrule
\end{tabular}
\label{tab:hyperparams_ebm}
\end{table}

\subsection{Non-context-free grammar parameterization }
\label{app:grammar-ncfg}

We consider a simple extension to the context-free grammar in which the expansion rules for a NT symbol can depend on its parent. We assume a product model structure for this dependence, i.e., if $P$ is a parent of $X$, then production probabilities from $X$ -- likelihoods of rules $X \to L\ R$ -- have a form
\[p_\theta(L,R\mid X,P)\propto f_1(L,R,X;\theta)f_2(L,R,P;\theta).\]
A context-free grammar corresponds to the case of $f_2$ being identically 1.

We note that a generative grammar of this form with $|NT|$ nonterminal symbols can be shown to equivalent to a context-free grammar with $|NT|^2$ nonterminal symbols by a standard construction. However, directly training grammars with, e.g., $30^2$ nonterminal symbols is prohibitive, while exact sampling from the posterior over parse trees in this non-context-free grammar has quintic time complexity (see \cref{tab:grammar-run-time}).

This intractability also makes calculating the marginal likelihood difficult.
As a result, we use a variational lower bound for that quantity by noting that
\begin{equation}
\begin{split}
    p_\theta(x) = \sum_{z}p_\theta(x, z) \ge \sum_{\tau\sim P_F; \tau\ni z}p_\theta(x, z) = F(s_0\mid x)
\end{split}
\end{equation}
where $F$ is the flow estimator.
Overall, we have $p_\theta(x) \ge F(s_0\mid x)$ when the GFlowNet has converged.
Thus, we use $F(s_0\mid x)$, i.e., the flow of the initial state, as an estimate of a lower bound of the marginal likelihood.

\subsection{Time complexity analysis}
\label{app:grammar-run-time}
Replacing a highly optimized algorithm used for exact learning of CFGs with an amortized posterior estimator, e.g., parametrized by a Transformer, inevitably increases the run-time cost.
However, there are asymptotic advantages to GFlowNet-EM compared to exact baselines, i.e., \textit{Marginalization} and \textit{Exact-sampling EM}.

\autoref{tab:grammar-run-time} compares their theoretical time complexity.
We use $n$ to denote the length of input sequences and $|S|$ the number of possible symbols in $z$.
This work focuses on presenting the method of GFlowNet-EM, and we do optimize for run time efficiency.
For example, we use a Transformer with full attention, which has a time complexity of $\mathcal{O}(n^2)$ per forward pass in sequence length, making the complexity per trajectory $\mathcal{O}(n^3)$, even though the theoretical complexity is just quadratic.
This can be solved by simply using an architecture like a Transformer with linear attention.

\begin{table}[t]
\centering
\caption{Theoretical time complexity of GFlowNet-EM and exact baselines for both the CFG and the Non-CFG we introduce as a function of $n$, the number of NT symbols, and $|S|$, the length of the terminal symbol sequence. GFlowNet-EM is more efficient asymptotically due to amortization. The exact baselines become intractable on the Non-CFG.}
\begin{tabular}{@{}llc}
\toprule
Grammar                  & Method             & Time complexity                 \\ \midrule
 \multirow{2}{*}{CFG}    & Marginalization    & \multirow{2}{*}{$\mathcal{O}(n^3|S|^3)$}  \\
                         & Exact-sampling EM  &        \\\cmidrule(lr){2-3}
                         & GFlowNet-EM        & $\mathcal{O}(n^2|S|)$       \\
\midrule
\multirow{2}{*}{Non-CFG} & Marginalization    & \multirow{2}{*}{$\mathcal{O}(n^5|S|^5)$}\\
                         & Exact-sampling EM  &                               \\\cmidrule(lr){2-3}
                         & GFlowNet-EM        & $\mathcal{O}(n^2|S|)$             \\ \bottomrule
\end{tabular}
\label{tab:grammar-run-time}
\end{table}

\iffalse
\begin{table}[t]
\centering
\caption{Theoretical time complexity of GFlowNet-EM and exact baselines for both the CFG and the Non-CFG we introduce as a function of $n$, the number of NT symbols, and $|S|$, the length of the terminal symbol sequence. GFlowNet-EM is more efficient asymptotically due to amortization. The exact baselines become intractable on the Non-CFG.}
%
\begin{tabular}{@{}llcc}
\toprule
Grammar                  & Method             & Time complexity                        &  Space complexity \\ \midrule
 \multirow{2}{*}{CFG}    & Marginalization    & \multirow{2}{*}{$\mathcal{O}(n^3|S|^3)$}  & \multirow{2}{*}{$\mathcal{O}(n^?|S|^?)$}\\
                         & Exact-sampling EM  &                                           &   \\\cmidrule(lr){2-4}
                         & GFlowNet-EM        & $\mathcal{O}(n^2|S|)$                   & $\mathcal{O}(n|S|)$ \\
\midrule
\multirow{2}{*}{Non-CFG} & Marginalization    & \multirow{2}{*}{$\mathcal{O}(n^5|S|^5)$} & \multirow{2}{*}{$\mathcal{O}(n^?|S|^?)$}\\
                         & Exact-sampling EM  &                                          &               \\\cmidrule(lr){2-4}
                         & GFlowNet-EM        & $\mathcal{O}(n^2|S|)$                  & $\mathcal{O}(n|S|)$   \\ \bottomrule
\end{tabular}
%
\label{tab:grammar-run-time}
\end{table}
\fi
%

\subsection{Ablation studies}
\label{app:grammar-ablation}

To understand the impact of the optimization techniques for GFlowNet-EM introduced in \autoref{sec:gfnem-tricks}, we perform three ablation studies.

\paragraph{Effect of E-step optimization techniques.} In the first study, we focus on the E-step. We fix a grammar learned with \textit{Marginalization} to compute the reward for the GFlowNet. As the metric of comparison we use an upper bound on the negative marginal log-likelihood per word under the GFlowNet, given by $\log \frac{p_\theta(x|z)p_\theta(z)P_B({s_0\rightarrow\dots\rightarrow z}|x)}{P_F({s_0\rightarrow\dots\rightarrow z}|x)}$. The results are summarized in \autoref{tab:fixed_grammar_ablation}. 

There are a few takeaways from this experiment. It is clear that a combination of all the optimization techniques is necessary for the best performance. Further, we also compare to a hierarchical VI (HVI) baseline due to its close connection with GFlowNets~\cite{malkin2022gfnhvi}. We observe that HVI performs the worst.

\paragraph{Joint learning.} To understand the full impact of the techniques, we consider the full joint learning scenario in \autoref{tab:joint_learning_ablation}. Again, we observe that all techniques are required to get the best performance. Notably, despite strong performance on the fixed grammar, only using the Sleep phase performs much worse in the case of joint learning. This is potentially due to the fact that reward in joint learning is non-stationary and thus hard to model without exploration.

In a separate experiment, we consider the of the threshold used in the adaptive E-step. The results are summarized in~\autoref{tab:threshold_ablation}. In summary, although thresholding is necessary -- without it, the generative model tends to collapse, i.e., most symbols of the grammar receive almost zero mass in the posterior -- if the threshold is sufficiently low, then its exact value controls the convergence rate, trading off between speed and accuracy of fitting the posterior. All thresholds result in convergence to similar final NLL and F1 scores, but we observe faster convergence to these values with higher values of the threshold.

\begin{table}[t]
\centering
\caption{Ablation on training a GFlowNet to sample from the posterior of a fixed grammar using a variational upper bound on the marginal NLL per word. All configurations are run with 5 random seeds.}
\begin{tabular}{cccc}
\toprule
GFlowNet loss & Exploration & Sleep & NLL / word $\downarrow$ \\\midrule 
SubTB & $\checkmark$ & $\checkmark$ & ${\bf \leq5.97\pm 0.01}$ \\
SubTB & $\checkmark$ & $\times$ & $\leq8.56\pm1.74$ \\
TB & $\checkmark$ & $\times$ & $\leq8.95\pm1.36$ \\
TB & $\times$ & $\times$ & $\leq9.40\pm1.44$ \\
HVI & $\times$ & $\times$ & $\leq13.14\pm1.55$ \\
$\times$ & $\times$ & $\checkmark$ & $\leq6.03 \pm 0.01$ \\\midrule
\multicolumn{3}{c}{Exact NLL / word} & 5.65 \\
\bottomrule
\end{tabular}
\label{tab:fixed_grammar_ablation}
\end{table}

\begin{table}[t]
\centering
\caption{Ablation on training GFlowNet-EM on a context-free grammar with different threshold schedules. We vary the threshold at the start and end of training, with linear decay in between. A higher threshold results in more frequent M-steps. All configurations are run with 5 random seeds.}
\begin{tabular}{lcc}
\toprule
Threshold & NLL / word $\downarrow$ & Sentence F1 $\uparrow$ \\\midrule 
$12 \to 6$ &	$5.77\pm0.02$	& $32.12\pm2.93$ \\
$8 \to 4$  &	$5.78\pm0.02$	& $32.23\pm3.25$ \\
$6 \to 3$  &	$5.76\pm0.02$	& $34.49\pm2.81$ \\
$4 \to 2$  &	$5.79\pm0.02$	& $30.56\pm4.42$ \\
\bottomrule
\end{tabular}
\label{tab:threshold_ablation}
\end{table}

\begin{table}[t]
\centering
\caption{Ablations on joint learning in GFlowNet-EM for CFG. All configurations are run over 5 random seeds.}
\begin{tabular}{@{}lcc}
\toprule
Method         & NLL / word $\downarrow$ &  Sentence F1 $\uparrow$ \\ \midrule
GFlowNet-EM    & ${\bf 5.70\pm 0.03}$ & ${\bf 34.85\pm3.39}$\\
\small{\ \ \ \ \ \ $-$MCMC}   & $6.02\pm 0.01$ &  $28.56\pm0.55$ \\ %
\small{\ \ \ \ \ \ $-$Sleep}    &  $5.91\pm 0.04$ & $28.13\pm 0.43$ \\
\small{\ \ \ \ \ \ $-$SubTB}    &   $5.84\pm 0.08$ &      $26.56\pm7.82$       \\
\small{\ \ \ \ \ \ $-$Exploration}   & ${\bf 5.70\pm0.02}$ &  $31.87\pm1.06$ \\ \midrule
\small{Sleep Only}  & $6.08\pm 0.06$ &  $48.41\pm 1.38$\\ \bottomrule
\end{tabular}
\label{tab:joint_learning_ablation}
\end{table}

\subsection{Sample parses from grammars learned by GFlowNet-EM}

GFlowNet-EM is able to learn diverse tree structures for both the CFG (without EBMs)~(\autoref{tab:sample_parses_cfg}) and the NCFG case~(\autoref{tab:sample_parses_ncfg}) but collapses to right-deep trees when guided by an EBM~(\autoref{tab:sample_parses_ebm}).
The learned latent structures that lead to better modeling of the data don't necessarily agree with our linguistic intuition.
The GFlowNet parser does not use diverse tags for top of the parse trees.
This may indicate either that high-level rules are harder to learn because they depend on meaningful low-level tags, or that a greater improvement in likelihood can be achieved by better hierarchical modeling of low-level structure (e.g., having latent tags responsible for frequent bigrams).
Both observations can motivate future work on more interpretable latents and better exploring the latent space using GFlowNet-EM.

\begin{table}[ht]
    \centering
    \caption{Sample parses generated with GFlowNet-EM on a context-free grammar.}
\resizebox{\textwidth}{0.5\textheight}{
    \begin{tabular}{ccc}\toprule
         Parse Tree & $\log Z(x)$ & $\log p(z|x)$ \\ \midrule
         \begin{tikzpicture}
\Tree
[.Q15 [.Q15 [.Q15 [.Q14 [portrait ] [studios ] ] ] [.Q7 [have ] [also ] ] ] [<unk> ] ] [.Q22 [onto ] [.Q12 [the ] [trend ] ] ] ]
      \end{tikzpicture}& $-43.68$  &  
$-49.19$ \\
    \begin{tikzpicture}
\Tree
[.Q15 [.Q15 but [.Q15 [.Q15 [.Q15 [.Q15 [.Q15 then [.Q4 as quickly ] ] [.Q3 as [.Q12 the dow ] ] ] [.Q22 had fallen ] ] [.Q23 it began ] ] [.Q18 to turn ] ] ] around ]
      \end{tikzpicture}& $-112.76$  &  
$-126.26$ \\
    \begin{tikzpicture}
\Tree
[.Q28 [.Q28 [.Q28 [.Q28 [.Q10 [.Q3 [\textit{to} ] [\textit{make} ] ] [\textit{them} ] ] [.Q13 [.Q6 [\textit{directly} ] [\textit{comparable} ] ] [.Q13 [.Q27 [\textit{each} ] [\textit{index} ] ] [.Q13 [\textit{is} ] [\textit{based} ] ] ] ] ] [.Q22 [\textit{on} ] [.Q11 [\textit{the} ] [\textit{close} ] ] ] ] [.Q9 [\textit{of} ] [\textit{N} ] ] ] [.Q9 [\textit{equaling} ] [\textit{N} ] ] ]
      \end{tikzpicture}& $-85.03$  &  
$-94.73$ \\\bottomrule
    \end{tabular}
}
    \label{tab:sample_parses_cfg}
\end{table}

\begin{table}[ht]
    \centering
    \caption{Sample parses generated with GFlowNet-EM on a context-free grammar with an annealed EBM prior.}
    \resizebox{\textwidth}{0.5\textheight}{
    \begin{tabular}{ccc}\toprule
         Parse & $\log Z(x)$ & $\log p(z|x)$ \\ \midrule
         \begin{tikzpicture}
\Tree[.Q3 [\textit{this} ] [.Q25 [\textit{market} ] [.Q14 [\textit{has} ] [.Q25 [\textit{been} ] [.Q25 [\textit{very} ] [.Q25 [\textit{badly} ] [\textit{damaged} ] ] ] ] ] ] ]
      \end{tikzpicture}& $-48.21$  &  
$-51.02$ \\
    \begin{tikzpicture}
\Tree[.Q3 [\textit{u.s.} ] [.Q25 [\textit{treasury} ] [.Q25 [\textit{bonds} ] [.Q14 [\textit{were} ] [.Q13 [\textit{higher} ] [.Q14 [\textit{in} ] [.Q3 [\textit{overnight} ] [.Q25 [\textit{trading} ] [.Q14 [\textit{in} ] [.Q3 [\textit{japan} ] [.Q14 [\textit{which} ] [.Q14 [\textit{opened} ] [.Q14 [\textit{at} ] [.Q3 [\textit{about} ] [.Q3 [\textit{N:N} ] [.Q25 [\textit{p.m.} ] [\textit{edt} ] ] ] ] ] ] ] ] ] ] ] ] ] ] ] ] ]
      \end{tikzpicture}& $-118.52$  &  
$-114.69$ \\
    \begin{tikzpicture}
\Tree[.Q3 [\textit{to} ] [.Q25 [\textit{make} ] [.Q3 [\textit{them} ] [.Q14 [\textit{directly} ] [.Q25 [\textit{comparable} ] [.Q3 [\textit{each} ] [.Q25 [\textit{index} ] [.Q14 [\textit{is} ] [.Q25 [\textit{based} ] [.Q14 [\textit{on} ] [.Q3 [\textit{the} ] [.Q25 [\textit{close} ] [.Q14 [\textit{of} ] [.Q23 [\textit{N} ] [.Q12 [\textit{equaling} ] [\textit{N} ] ] ] ] ] ] ] ] ] ] ] ] ] ] ] ]
      \end{tikzpicture}& $-94.38$  &  
$-96.89$ \\\bottomrule
    \end{tabular}
    }
    \label{tab:sample_parses_ebm}
\end{table}

\begin{table}[ht]
    \centering
    \caption{Sample parses generated with GFlowNet-EM on a non-context-free grammar.}
\resizebox{\textwidth}{0.5\textheight}{
    \begin{tabular}{ccc}\toprule
         Parse & $\log Z(x)$ & $\log p(z|x)$ \\ \midrule
         \begin{tikzpicture}
\Tree
[.Q9 [portrait ] [.Q9 [studios ] [.Q9 [.Q25 [.Q25 [have ] [also ] ] [<unk> ] ] [.Q5 [onto ] [.Q11 [the ] [trend ] ] ] ] ] ]
      \end{tikzpicture}& $-42.11$  &  
$-50.74$ \\
    \begin{tikzpicture}
\Tree
[.Q9 [.Q11 [but ] [then ] ] [.Q9 [.Q16 [as ] [quickly ] ] [.Q9 [.Q25 [as ] [.Q11 [the ] [dow ] ] ] [.Q5 [.Q25 [had ] [fallen ] ] [.Q5 [.Q25 [it ] [.Q28 [.Q10 [began ] [to ] ] [turn ] ] ] [around ] ] ] ] ] ]
      \end{tikzpicture}& $-117.60$  &  
$-153.98$ \\
    \begin{tikzpicture}
\Tree
[.Q17 [.Q7 [.Q27 [\textit{to} ] [\textit{make} ] ] [\textit{them} ] ] [.Q17 [.Q3 [\textit{directly} ] [.Q12 [\textit{comparable} ] [\textit{each} ] ] ] [.Q17 [\textit{index} ] [.Q17 [.Q15 [.Q15 [\textit{is} ] [\textit{based} ] ] [.Q10 [.Q12 [\textit{on} ] [.Q7 [\textit{the} ] [\textit{close} ] ] ] [.Q20 [\textit{of} ] [\textit{N} ] ] ] ] [.Q10 [\textit{equaling} ] [\textit{N} ] ] ] ] ] ]
      \end{tikzpicture}& $-91.77$  &  
$-105.96$ \\\bottomrule
    \end{tabular}
}
    \label{tab:sample_parses_ncfg}
\end{table}

\section{Discrete VAE}
\label{app:vae}

\subsection{Experiment setup}
In all our experiments we use a $4\times4$ discrete latent representation and increase the number of categorical entries $K$ in the dictionary from 4 to 8 to 10. We omitted larger dictionary sizes as we observed the VQ-VAE NLL worsen for $K\geq10$. In total there are $K^16$ possible latent configurations.

We used a similar architecture as the one described in \cite{oord2017neural}, adding batch normalization and additional downsizing and upsizing convolutional layers to obtain the smaller $4\times4$ latent representation. The GFlowNet encoder network extends the VQ-VAE convolutional image encoder by adding state encoding and state prediction MLPs. The decoder network is the same for both models. The prior distribution is modeled using a PixelCNN \cite{oord2016pixel} with 8 masked convolution layers.

For $K=\{4,8\}$ we trained the VQ-VAE model for 50 epochs with a learning rate of $2 \times 10^{-4}$, reduced to $5 \times 10^{-5}$ at epoch 25. For $K=10$, we trained for 80 epochs with the same learning rate, which was now reduced at epoch 50. The GFlowNet-EM + Greedy Decoder model was trained in all settings for 250 epochs, with a learning rate of $2 \times 10^{-4}$, reduced to $5 \times 10^{-5}$ at epoch 180. For the experiments where the decoder is not trained with greedily-drawn samples, the training steps were doubled. Lastly, in the GFlowNet-EM + Prior experiments the models were trained trained for 400 epochs with similar learning rate schedules, and the reduction at epoch 300. We clarify that each GFlowNet-EM epoch is inherently slower since the encoder network requires multiple forward passes to construct the latent representation. We disabled all batch normalization layers for the GFlowNet experiments and used a batch size of 128 in all our tests. 

\subsection{GFlowNet-EM visualizations}
\begin{figure}[t]
\centering
\includegraphics[width=0.99\linewidth,trim=0 0 0 0]{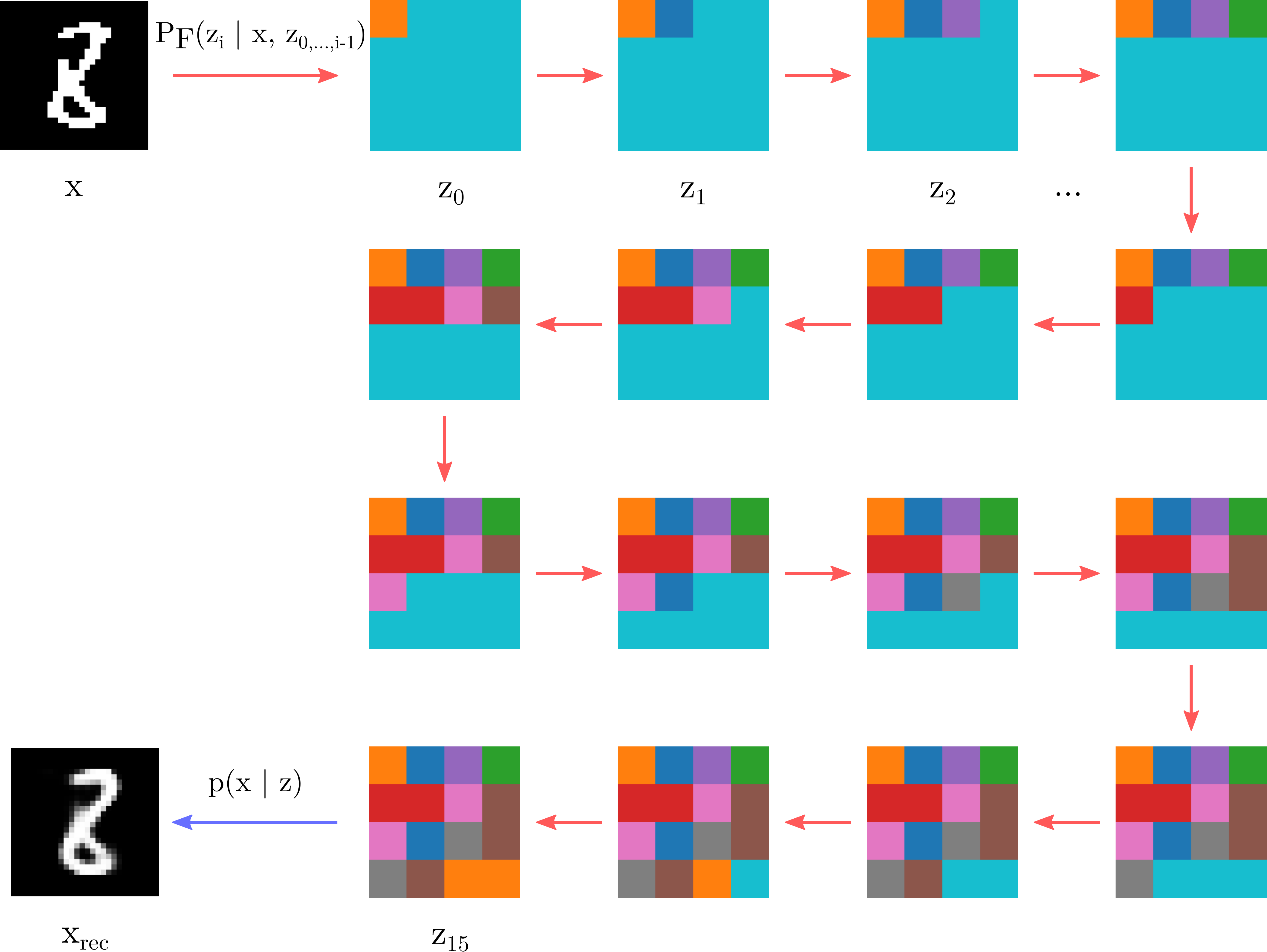}
\caption{Visualization of the procedure of encoding an image $x$ into a discrete representation $z$ using a GFlowNet encoder with an autoregressive policy and reconstructing the original image.}
 \label{fig:vae_policy}
\end{figure}

\begin{figure}[t]
\centering
\begin{tabular}{cccc}
\includegraphics[width=0.12\linewidth,trim=0 0 0 0]{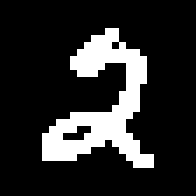}
&
\includegraphics[width=0.12\linewidth,trim=0 0 0 0]{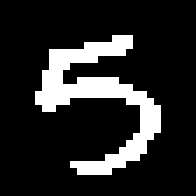}
&
\includegraphics[width=0.12\linewidth,trim=0 0 0 0]{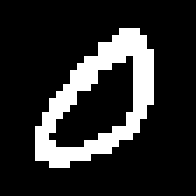}
&
\includegraphics[width=0.12\linewidth,trim=0 0 0 0]{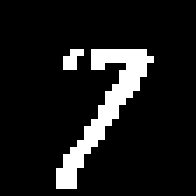}
\\
\includegraphics[width=0.12\linewidth,trim=0 0 0 0]{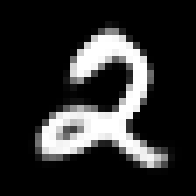}
&
\includegraphics[width=0.12\linewidth,trim=0 0 0 0]{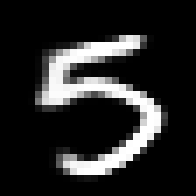}
&
\includegraphics[width=0.12\linewidth,trim=0 0 0 0]{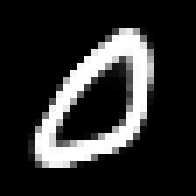}
&
\includegraphics[width=0.12\linewidth,trim=0 0 0 0]{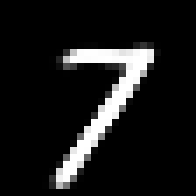}
\end{tabular}
\caption{Images from the static MNIST test set and their reconstructions using the GFlowNet-EM model with $K=8$.}
 \label{fig:vae_reconstructions}
\end{figure}

\begin{figure}[t]
\centering
\begin{tabular}{ccc}
\includegraphics[width=0.16\linewidth,trim=0 0 0 0]{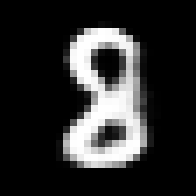}
&
\includegraphics[width=0.16\linewidth,trim=0 0 0 0]{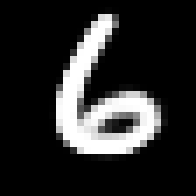}
&
\includegraphics[width=0.16\linewidth,trim=0 0 0 0]{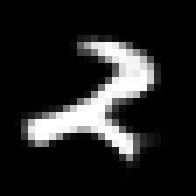}
\\
\includegraphics[width=0.16\linewidth,trim=0 0 0 0]{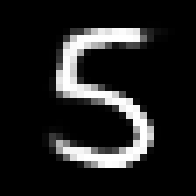}
&
\includegraphics[width=0.16\linewidth,trim=0 0 0 0]{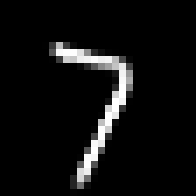}
&
\includegraphics[width=0.16\linewidth,trim=0 0 0 0]{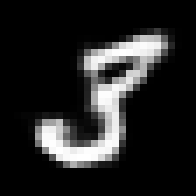}
\\
\includegraphics[width=0.16\linewidth,trim=0 0 0 0]{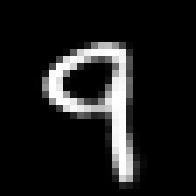}
&
\includegraphics[width=0.16\linewidth,trim=0 0 0 0]{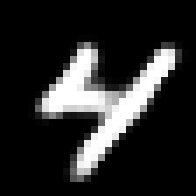}
&
\includegraphics[width=0.16\linewidth,trim=0 0 0 0]{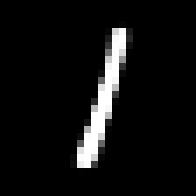}
\end{tabular}
\caption{Samples drawn from the learned prior of the GFlowNet-EM model with dictionary size $K=8$.}
 \label{fig:vae_prior_samples}
\end{figure}

In Fig.~\ref{fig:vae_policy} we visualize the steps of encoding an input image into a discrete representation and reconstructing it. We limited the GFlowNet policy to be autoregressive, which we found to strike an appropriate balance between posterior expressiveness and model complexity. At every step the GFlowNet encoder 'looks' at the image and existing state and samples the next entry in the latent representation. In Figures~\ref{fig:vae_reconstructions} and \ref{fig:vae_prior_samples} we present results of the GFlowNet-EM model with dictionary size $K=8$ and a jointly learned prior. Despite the minimal limited latent representation, the model has captured the variety in the data which we showcase in the samples drawn from the learned prior.

\section{Computation cost in practice}
\label{app:computation_cost}

\paragraph{Grammar induction}
Our experiments with the context-free grammar take 23 hours to run to completion on a single V100 GPU, while the baseline from~\citet{kim-etal-2019-compound} takes 21 hours to run on similar hardware. For completeness' sake, we note that a specialized library called torch-struct was later developed on the basis of~\citet{kim-etal-2019-compound}'s work, introducing several optimization tricks that reduce the computation time 8-fold. We can expect software optimizations to similarly help speed up GFlowNet-EM.

However, in the non-context-free case, a GFlowNet-EM run still takes roughly 23 hours, while exact parsing using the (generalization of) inside algorithm will take orders of magnitude longer in the absence of conditional independence assumptions. With the EBM prior on tree shape, exact parsing is completely intractable (no longer even polynomial in sequence length).

\paragraph{Discrete VAE}
We used a fixed number of updates for both the E and M steps. An E-step (training the GFlowNet encoder) takes approximately 25s for 400 updates, whereas the M-step (training the convolutional decoder) requires 10s for 400 updates on one A5000 GPU. In the base experiment, training takes rougly 3 hours, which is halved when the greedy decoder is used. When also learning the prior, the E and M steps take 26s and 18s respectively. Training requires again approximately 3 hours. In comparison, training any of the VQ-VAE models requires about 15m, indicating a overhead for GFlowNet-EM with greedy decoder training. Future work should consider ways to accelerate GFlowNet-EM training, such as by better selection of learning rates and update schedules for the E, sleep, and M steps, which we did not extensively tune for the experiments in this paper.

\end{document}